\pdfoutput=1

\documentclass[11pt]{article}

\newif\ifarxiv
\arxivtrue

\ifarxiv
    \usepackage[final]{acl}
\else
    \usepackage[review]{acl}
\fi

\usepackage{times}
\usepackage{latexsym}
\usepackage{amsmath,amssymb,amsthm}
\usepackage{bbm}
\usepackage{mathabx,mathrsfs}
\usepackage{relsize}
\PassOptionsToPackage{hyphens}{url}\usepackage{hyperref}
\usepackage{xurl}
\usepackage[T1]{fontenc}

\usepackage{CJKutf8}
\usepackage[utf8]{inputenc}

\usepackage{microtype}

\usepackage{inconsolata}

\usepackage{natbib}

\usepackage{todonotes}
\usepackage{microtype}
\usepackage{pifont}
\usepackage{bm}
\usepackage{amsmath}
\usepackage{amssymb}
\usepackage{autobreak}
\usepackage{float}
\usepackage{subcaption}
\usepackage{paralist}
\usepackage{xpinyin}
\usepackage{scalerel}
\usepackage{tikz}
\usepackage{tikz-dependency}
\usepackage{tikz-qtree}
\usepackage{tikz-layers}
\usepackage{tikzsymbols}
\usepackage{pgfplots}
\usepackage{graphicx}
\usepackage{adjustbox}
\usepgflibrary{arrows.meta}
\usepgfplotslibrary{fillbetween, statistics, groupplots}
\usetikzlibrary{fit, calc, plotmarks, decorations.pathreplacing, decorations.markings, positioning, arrows.meta, shapes.geometric, backgrounds, graphs}
\pgfplotsset{compat=1.18}
\usepackage{siunitx}
\usepackage{rotating}
\usepackage{booktabs}
\usepackage{tcolorbox}
\usepackage{xcolor,colortbl}
\usepackage{nicematrix}
\usepackage{mleftright}
\usepackage{multirow}

\newcommand\Letter{{\fontfamily{mvs}\fontencoding{U}\selectfont\char66}}

\def\myemoji@insert#1{\scalerel*{\includegraphics[page=#1]{assets/myemoji-assets.pdf}}{X}}
\newcommand\orange{1}
\newcommand\coconut{2}
\def\hwemoji@insert#1{\scalerel*{\includegraphics[page=#1]{assets/hwemoji-assets.pdf}}{X}}
\newcommand\mailemoji{\hwemoji@insert{3610}}

\newcommand\snowflakeemoji{\hwemoji@insert{3644}}

\definecolor{json_blue}{RGB}{15, 89, 164}
\definecolor{json_red}{RGB}{192, 44, 53}
\definecolor{figure_green}{RGB}{32, 137, 77}
\definecolor{figure_blue}{RGB}{52, 108, 156}
\definecolor{figure_red}{RGB}{192, 44, 53}
\definecolor{figure_orange}{RGB}{250, 126, 35}
\definecolor{table_blue}{RGB}{92, 179, 204}
\definecolor{table_red}{RGB}{238, 63, 77}
\definecolor{table_orange}{RGB}{250, 126, 35}
\definecolor{figure_light_green}{RGB}{198, 223, 200}
\definecolor{figure_light_blue}{RGB}{208, 223, 230}
\definecolor{figure_light_red}{RGB}{192, 44, 53}
\definecolor{figure_light_gray}{RGB}{220, 220, 220}
\definecolor{figure_gray}{RGB}{160, 160, 160}

\newcommand\sideline[1]{\texttt{{\textcolor{gray}{#1}}}}

\newcommand\wz{\phantom{0}}
\newcommand\wm{\phantom{\texttt{-}}}
\newcommand\ewm{\texttt{-}}
\newcommand\ewp{\texttt{+}}
\newcommand\bgood{\hspace{-0.05em}\rlap{\scalebox{0.9}{$^\upharpoonright$}}}
\newcommand\sgood{\hspace{-0.15em}\rlap{\scalebox{0.9}{$^\downharpoonright$}}}
\newcommand\correct[1]{\textcolor{figure_blue}{{#1}}}
\newcommand\wrong[1]{\textcolor{figure_red}{{{#1}}}}
\newcommand\ood[1]{\textcolor{figure_gray}{\textit{#1}}}

\usepackage{contour}
\usepackage[normalem]{ulem}
\usepackage{soul}

\setuldepth{Berlin}%

\NiceMatrixOptions
       {
         custom-line =
           {
             letter = ; ,
             command = hdashedline ,
             ccommand = cdashedline ,
             tikz = dashed
    } 
}

\pgfdeclarelayer{background}  %
\pgfdeclarelayer{above}       %
\pgfsetlayers{background,main,above}  %

\newcommand \footnotetextonly[1]
{
    \let \backupfootnote \thefootnote
    \let \thefootnote \relax
    \footnotetext{#1}
    \let \thefootnote \backupfootnote
    \let \backupfootnote \imreallyundefinedcommand
}

\title{A Simple yet Effective Training-free Prompt-free Approach\\to Chinese Spelling Correction Based on Large Language Models}

\ifarxiv
    \author{Houquan Zhou\rlap{$^{\orange}$},\ \ Zhenghua Li\rlap{$^{\orange\text{\Letter}}$},\ \ \ \ Bo Zhang\rlap{$^{\coconut}$},\ \ Chen Li\rlap{$^{\coconut}$} \\
    {\bf  Shaopeng Lai\rlap{$^{\coconut}$},\ \  Ji Zhang\rlap{$^{\coconut}$},\ \  Fei Huang\rlap{$^{\coconut}$},\ \  Min Zhang$^{\orange}$}  \\%
    $^{\orange}$School of Computer Science and Technology, 
    Soochow University, China \\%
    \texttt{hqzhou@stu.suda.edu.cn}, \texttt{\{zhli13,minzhang\}@suda.edu.cn}\\%
    $^{\coconut}$DAMO Academy, Alibaba Group, China\\%
    \texttt{\{klayzhang.zb,puji.lc,laishaopeng.lsp,zj122146,f.huang\}@alibaba-inc.com}}
\else
    \author{Anonymous}
\fi

\pgfplotsset{compat=1.17}
\begin{document}
\begin{CJK}{UTF8}{gkai}
    \maketitle
    \ifarxiv%
        \footnotetextonly{\!\!\Letter\ Zhenghua Li is the corresponding author.}
    \fi%

    \begin{abstract}
    This work proposes a simple training-free prompt-free approach to leverage large language models (LLMs) for the Chinese spelling correction (CSC) task, which is totally different from all previous CSC approaches.
    The key idea is to use an LLM as a pure language model in a conventional manner.
    The LLM goes through the input sentence from the beginning, and at each inference step, produces a distribution over its vocabulary for deciding the next token, given a partial sentence.
    To ensure that the output sentence remains faithful to the input sentence, we design a minimal distortion model that utilizes pronunciation or shape similarities between the original and replaced characters.
    Furthermore, we propose two useful reward strategies to address practical challenges specific to the CSC task.
    Experiments on five public datasets demonstrate that our approach significantly improves LLM performance, enabling them to compete with state-of-the-art domain-general CSC models.
\end{abstract}

    \section{Introduction}
Given a Chinese character, there may exist many others with the same or similar pronunciations, or with similar shapes.
This similarity can lead to incorrect character selection when using certain keyboard input methods.
It is worth noting that nowadays most Chinese users rely on Pinyin-based input methods.
Besides, optical character recognition (OCR) and automatic speech recognition (ASR) systems may also introduce errors during image/speech-to-text conversion.
Such incorrect characters in texts degrade communication efficiency, and sometimes even lead to misunderstanding.

\begin{figure}[tb!]
    \centering
    \scalebox{1.0}{
        \begin{tikzpicture}[
                font=\scriptsize,
                anchor point/.style={
                        draw=none,
                        circle,
                        inner sep=1.5pt,
                    },
                connect/.style={
                        shorten >= 2pt,
                        shorten <= 1.5pt,
                        thin,
                        rounded corners=1.5pt,
                        draw=black!60,
                    },
                arrow/.style={
                        connect,
                        arrows = {-Straight Barb[length=0.8mm]},
                    },
                predict arrow/.style={
                        arrow,
                        rounded corners=2.5pt,
                    },
                word/.style={
                        font=\footnotesize,
                        anchor=base,
                        inner sep=0pt,
                    },
                candidate/.style={
                        font=\scriptsize,
                        anchor=base,
                        draw=none,
                        inner sep=1pt,
                    },
                candidate placeholder/.style={
                        font=\footnotesize,
                        anchor=base,
                        draw=none,
                        minimum height=0.6em,
                        minimum width=2.1em,
                        inner sep=0pt,
                    },
                prob/.style={
                        font=\tiny,
                        anchor=base,
                        draw=none,
                        inner sep=0pt,
                    },
                label/.style={
                        font=\tiny,
                        draw=none,
                        inner sep=0pt,
                    },
                type/.style={
                        font=\tiny,
                        draw=none,
                        inner sep=1pt,
                    },
                union/.style={
                        fill=white,
                        draw=none,
                        inner sep=0pt,
                        thin
                    },
                candidate union/.style={
                        union,
                        semithick,
                        draw=figure_gray,
                        fill=figure_light_gray!20,
                        rounded corners=1.5pt,
                        inner sep=4pt,
                    },
                sentence union/.style={
                        union,
                        draw=figure_gray,
                        fill=figure_light_gray!10,
                        rounded corners=1.5pt,
                        inner sep=1.0pt,
                    },
                etc/.style={
                        draw=none,
                        font=\tiny,
                        inner sep=0pt,
                    },
            ]

            \graph [
            empty/.style={draw=none},
            nodes={anchor=base, inner sep=1pt},
            edges={arrows = {-Straight Barb[length=0.8mm]}},
            grow right sep=0.06cm,
            branch down=0.7,
            multi] {
            bos/"\sideline{\texttt{<BOS>}}" -!- y0/"明" -!- y1/"天" -!- y2/"\correct{就}" -!- y3/"\correct{是}" -!- y4/"周" -!- y5/"末" -!- y6/"了" -!- y7/"，\strut" [yshift=-0.08cm] -!- y8/"又" -!- y9/"可" -!- y10/"以";
            };
            \node[label, anchor=south] (y2_pinyin) at ($(y2.north) + (0, +0.02)$) {\correct{\textit{jiù\strut}}};
            \node[label, anchor=north] (y2_3_gloss) at ($(y2.south)!0.5!(y3.south) + (0, -0.05)$) {\correct{\textit{is\strut}}};
            \node[label, anchor=south] (y3_pinyin) at ($(y3.north) + (0, +0.02)$) {\correct{\textit{shì\strut}}};

            \begin{pgfonlayer}{background}
                \node[sentence union, fit=(bos) (y0) (y10)] (generated) {};
            \end{pgfonlayer}

            \node[anchor=south west, inner sep=0pt] (generated_label) at ($(generated.north west) + (0, 0.25)$) {\textbf{Partially Generated Sentence:}};

            \begin{scope}[yshift=-6cm]
                \graph [
                empty/.style={draw=none},
                nodes={anchor=base, inner sep=1pt},
                edges={arrows = {-Straight Barb[length=0.8mm]}},
                grow right sep=0.06cm,
                branch down=0.7,
                multi] {
                x0/"明" -!- x1/"天" -!- x2/"\wrong{九}" -!- x3/"\wrong{十}" -!- x4/"周" -!- x5/"末" -!- x6/"了" -!- x7/"，\strut" [yshift=-0.08cm] -!- x8/"又" -!- x9/"可" -!- x10/"以" -!- x11/"\wrong{根}" -!- x12/"朋" -!- x13/"友" -!- x14/"出" -!- x15/"去" -!- x16/"玩" -!- x17/"了" -!- x18/"。" [yshift=-0.08cm];
                };
            \end{scope}
            \node[label, anchor=south] (x2_pinyin) at ($(x2.north) + (0, -0.05)$) {\wrong{\textit{jiǔ\strut}}};
            \node[label, anchor=south] (x3_pinyin) at ($(x3.north) + (0, -0.05)$) {\wrong{\textit{shí\strut}}};
            \node[label, anchor=south] (x11_pinyin) at ($(x11.north) + (0, -0.05)$) {\wrong{\textit{gēn\strut}}};
            \node[label, anchor=south] (x12_pinyin) at ($(x12.north) + (0, -0.05)$) {\textit{péng\strut}};
            \node[label, anchor=south] (x13_pinyin) at ($(x13.north) + (0, -0.05)$) {\textit{yǒu\strut}};
            \node[label, anchor=south] (x14_pinyin) at ($(x14.north) + (0, -0.05)$) {\textit{chū\strut}};
            \node[label, anchor=south] (x15_pinyin) at ($(x15.north) + (0, -0.05)$) {\textit{qù\strut}};
            \node[label, anchor=south] (x16_pinyin) at ($(x16.north) + (0, -0.05)$) {\textit{wán\strut}};
            \node[label, anchor=south] (x17_pinyin) at ($(x17.north) + (0, -0.05)$) {\textit{le\strut}};
            \node[label, anchor=north] (x2_3_gloss) at ($(x2.south)!0.5!(x3.south) + (0, -0.01)$) {\wrong{\textit{ninety\strut}}};
            \node[label, anchor=north] (x11_gloss) at ($(x11.south) + (0, -0.01)$) {\wrong{\textit{root\strut}}};

            \node[anchor=south west, inner sep=0pt] (input_label) at ($(x0.north west-|generated.west) + (0, 0.2)$) {\textbf{Input Sentence:}};

            \begin{pgfonlayer}{background}
                \node[sentence union, fit=(x11) (x17_pinyin) (x18)] (input) {};
            \end{pgfonlayer}

            \coordinate (center) at ($(y0)!0.5!(x18) + (0, 0.5cm)$);

            \node[anchor point] (p0) at ($(center) + (-3.15, 0)$) {};
            \node[anchor point] (p1) at ($(center) + (-1.9, 0)$) {};
            \node[anchor point] (p2) at ($(center) + (-0.95, 0)$) {};
            \node[anchor point] (p3) at ($(center) + (0, 0)$) {};
            \node[anchor point] (p4) at ($(center) + (1.0, 0)$) {};
            \node[anchor point] (p5) at ($(center) + (2.1, 0)$) {};
            \node[anchor point] (p6) at ($(center) + (3.15, 0)$) {};

            \node[word] (c0) at (p0) {休息};
            \node[label, anchor=south] (c0_pinyin) at ($(c0.north) + (0, 0.05)$) {\textit{xiū xī\strut}};
            \node[label, anchor=south] (c0_placeholder) at ($(c0.north) + (0, 0.05)$) {\phantom{\textit{gh}}};
            \node[label, anchor=north] (c0_gloss) at ($(c0.south) + (0, -0.1)$) {\textit{rest\strut}};
            \node[prob, anchor=north] (c0_prob) at ($(c0_gloss.south) + (0, -0.05)$) {$0.162$};

            \node[word] (c1) at (p1) {睡};
            \node[label, anchor=south] (c1_pinyin) at ($(c1.north) + (0, 0.05)$) {\textit{shuì\strut}};
            \node[label, anchor=south] (c1_placeholder) at ($(c1.north) + (0, 0.05)$) {\phantom{\textit{gh}}};
            \node[label, anchor=north] (c1_gloss) at ($(c1.south) + (0, -0.1)$) {\textit{sleep\strut}};
            \node[prob, anchor=north] (c1_prob) at ($(c1_gloss.south) + (0, -0.05)$) {$0.079$};

            \node[word] (c2) at (p3) {跟};
            \node[label, anchor=south] (c2_pinyin) at ($(c2.north) + (0, 0.05)$) {\textit{gēn\strut}};
            \node[label, anchor=south] (c2_placeholder) at ($(c2.north) + (0, 0.05)$) {\phantom{\textit{gh}}};
            \node[label, anchor=north] (c2_gloss) at ($(c2.south) + (0, -0.1)$) {\textit{with\strut}};
            \node[prob, anchor=north] (c2_prob) at ($(c2_gloss.south) + (0, -0.05)$) {$0.0168$};

            \node[word] (c3) at (p5) {根};
            \node[label, anchor=south] (c3_pinyin) at ($(c3.north) + (0, 0.05)$)  {\textit{gēn\strut}};
            \node[label, anchor=south] (c3_placeholder) at ($(c3.north) + (0, 0.05)$) {\phantom{\textit{gh}}};
            \node[label, anchor=north] (c3_gloss) at ($(c3.south) + (0, -0.1)$) {\textit{root\strut}};
            \node[prob, anchor=north] (c3_prob) at ($(c3_gloss.south) + (0, -0.05)$) {$0.000002$};

            \begin{pgfonlayer}{background}
                \node[candidate union, densely dashed, fit=(c0) (c0_prob) (c0_placeholder)] (c0_union) {};
                \node[candidate union, densely dashed, fit=(c1) (c1_prob) (c1_placeholder)] (c1_union) {};
                \node[candidate union, fit=(c2) (c2_prob) (c2_placeholder), fill=figure_light_blue, thick, draw=black] (c2_union) {};
                \node[candidate union, fit=(c3) (c3_prob) (c3_placeholder), fill=figure_light_blue!20, semithick] (c3_union) {};
            \end{pgfonlayer}
            \node[label, anchor=base east, inner sep=2pt] at (c2_prob.base-|c2_union.west) {$p_{\mathtt{LLM}}$:};

            \node[etc] at ($(c1_union.east)!0.5!(c2_union.west)$) {$\cdots$};
            \node[etc] at ($(c2_union.east)!0.5!(c3_union.west)$) {$\cdots$};
            \node[etc] at (p6 |- c0_union) {$\cdots$};

            \node[label, anchor=north, inner sep=1pt, align=center] (c2_type) at ($(c2_union.south) + (0, -0.1)$) {\textbf{\texttt{Same}}\\\textbf{\texttt{Pinyin}}};
            \node[label, inner sep=1pt] (c0_type) at (c0_union.south |- c2_type) {\texttt{Unrelated}};
            \node[label, inner sep=1pt] (c1_type) at (c1_union.south |- c2_type) {\texttt{Unrelated}};
            \node[label, inner sep=1pt] (c3_type) at (c3_union.south |- c2_type) {\textbf{\texttt{Identical}}};

            \node[prob, anchor=north, inner sep=1pt, rounded corners=1.0pt, fill=figure_light_blue!40] (c2_type_prob) at (c2_type.south) {$0.023$};
            \node[prob, inner sep=-1pt] (forbid_c0) at (c2_type_prob.base-|c0_type) {\scalerel*{\includegraphics{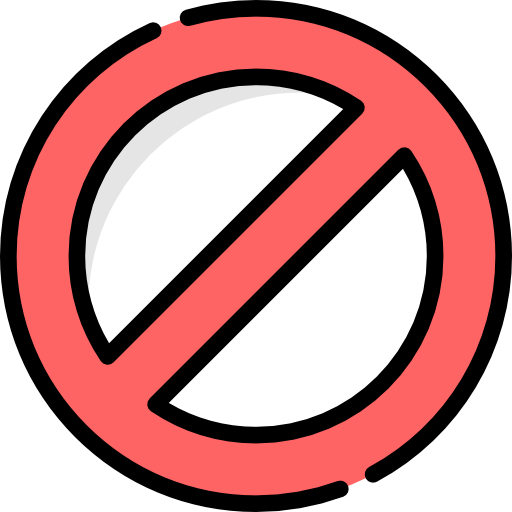}}{\strut}};
            \node[prob, inner sep=-1pt] (forbid_c1) at (c2_type_prob.base-|c1_type) {\scalerel*{\includegraphics{assets/forbid.png}}{\strut}};
            \node[prob, inner sep=1pt, rounded corners=1.0pt, fill=figure_light_blue!90] (c3_type_prob) at (c2_type_prob.base-|c3_type) {$0.962$};
            \node[label, anchor=base east, inner sep=2pt] at (c2_type_prob.base west) {$p_{\mathtt{DM}}$:};

            \node[anchor=center, draw=none, inner sep=0pt] (robot) at ($(center|-generated) + (0, -0.8)$){\includegraphics[width=0.6cm]{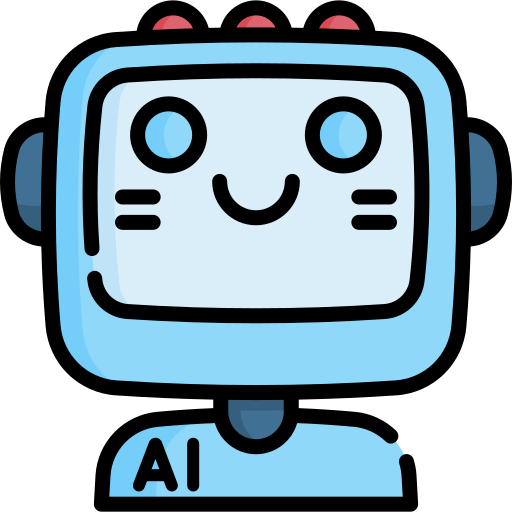}};
            \node[anchor=south east, draw=black, fill=white, inner sep=0.5pt, rounded corners=1.0pt, thin] (snowflake) at (robot.south east) {\snowflakeemoji};
            \node[anchor=east, inner sep=0pt] (robot_label) at ($(robot.west) + (-0.1, -0.15)$) {\textbf{Large Language Model}};
            \node[label, anchor=west, inner sep=0pt] at ($(c0_union.north west-|generated.west) + (0, +0.58)$) {\texttt{Next Token Prediction}};

            \node[anchor=center, draw=none, inner sep=0pt] (distortion) at ($(center|-input) + (0, 0.7)$){\includegraphics[width=0.6cm]{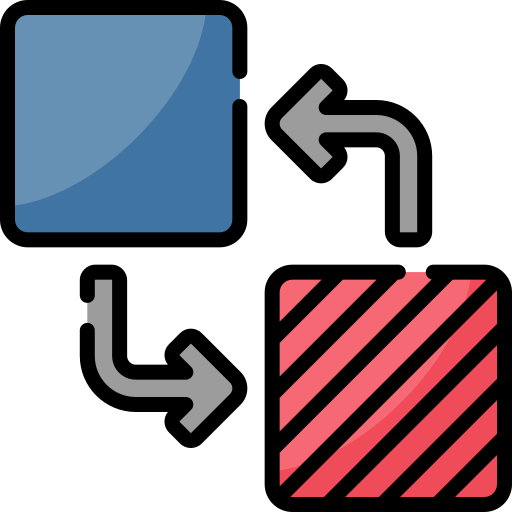}};
            \node[anchor=east, inner sep=0pt] (distortion_label) at ($(distortion.west) + (-0.1, 0.1)$) {\textbf{Minimal Distortion Model}};
            \node[label, anchor=west, inner sep=0pt] at ($(distortion.north-|generated.west) + (0, +0.18)$) {\texttt{Relationship Evaluation}};

            \draw[arrow] (generated) -- ++(0, -0.4) -- (robot);
            \draw[predict arrow] (robot) -- ++(0, -0.5) -- ($(c0_union.north) + (0, 0.3)$) -- (c0_union.north);
            \draw[predict arrow] (robot) -- ++(0, -0.5) -- ($(c1_union.north) + (0, 0.3)$) -- (c1_union.north);
            \draw[predict arrow] (robot) -- ++(0, -0.5) -- ($(c2_union.north) + (0, 0.3)$) -- (c2_union.north);
            \draw[predict arrow] (robot) -- ++(0, -0.5) -- ($(c3_union.north) + (0, 0.3)$) -- (c3_union.north);

            \draw[predict arrow] (distortion) -- ($(input) + (0, 0.6)$) -- (input);
            \draw[predict arrow] (distortion) -- ++(0, 0.5) -- ($(forbid_c0.south) + (0, -0.3)$) -- (forbid_c0.south);
            \draw[predict arrow] (distortion) -- ++(0, 0.5) -- ($(forbid_c1.south) + (0, -0.3)$) -- (forbid_c1.south);
            \draw[predict arrow] (distortion) -- ++(0, 0.5) -- ($(c2_type_prob.south) + (0, -0.3)$) -- (c2_type_prob.south);
            \draw[predict arrow] (distortion) -- ++(0, 0.5) -- ($(c3_type_prob.south) + (0, -0.3)$) -- (c3_type_prob.south);

            \node[prob, anchor=west, inner sep=1pt, rounded corners=1.0pt, fill=figure_light_blue] (final_prob_c2) at ($(c2_type.east) + (0.1, 0)$) {$0.00039$};
            \node[prob, anchor=west, inner sep=1pt, rounded corners=1.0pt, fill=figure_light_blue!30] (final_prob_c3) at ($(c3_type.east) + (0.0, 0)$) {$1.9\text{E-}6$};

            \draw[connect, draw=black, shorten >= 2.5pt, shorten <= 2.5pt] (c2_union) -- (final_prob_c2);
            \draw[connect, draw=black, shorten >= 2.5pt, shorten <= 2.5pt] (c2_type_prob.east) -- (final_prob_c2.south);
            \node[prob, figure_blue, anchor=south, font=\footnotesize] at ($(final_prob_c2.north) + (-0.15, +0.44)$) {\ding{52}};

            \draw[connect, draw=black, shorten >= 2.5pt, shorten <= 2.5pt] (c3_union) -- (final_prob_c3);
            \draw[connect, draw=black, shorten >= 2.5pt, shorten <= 2.5pt] (c3_type_prob.east) -- (final_prob_c3.south);

            \draw[predict arrow, rounded corners=8pt, draw=figure_blue, densely dashed, semithick] (c2_union) -- ++(1.4, 1.1) -- node[above, type, figure_blue, fill=white] {\texttt{Append}: 跟} ($(generated -| c2_union) + (1.4, -0.2)$) -- (generated.east);

        \end{tikzpicture}
    }
    \caption{
        An illustration of our approach.
        The correct sentence should be ``明天\correct{就是}周末了，又可以\correct{跟}朋友出去玩了。'' (\textit{Tomorrow \correct{is} the weekend, allowing for going out to play \correct{with} friends again.}).
    }
    \label{fig:illustration}
\end{figure}

As illustrated in Figure \ref{fig:illustration}, the task of Chinese spelling correction (CSC) aims to correct each incorrect character in a sentence \cite{yu-li-2014-chinese}, and has attracted a lot of attention in recent years \cite{bao-etal-2020-chunk,xu-etal-2021-read,li-etal-2022-improving-chinese,wu-etal-2023-rethinking,dong-etal-2024-rich}.

Recently, witnessing the success of large language models (LLMs), researchers try to leverage LLMs for the CSC task.
These approaches fall into two categories: \emph{prompt-based} and \emph{supervised fine-tuning} (SFT).
However, their performance lags behind non-LLM approaches by large margin.

The prompt-based approach relies on carefully designed prompts, using instructions with no example (zero-shot) or a few examples (few-shot), and requires a capable LLM (typically ChatGPT) to perform CSC \cite{li-etal-2023-ineffectiveness,dong-etal-2024-rich}.
However, spelling errors can sometimes make it very difficult for the LLM to correctly understand the original meaning of the sentence.
As a result, the LLM either ignores errors or replace an erroneous character to another erroneous one.
Moreover, extra strategies are required to help ChatGPT ensure that the length of the output sentence is consistent with the input sentence.

The SFT approach also uses prompts but, in contrast to the prompt-based approach, it continues tuning the parameters of an LLM using CSC training data \cite{li-etal-2023-ineffectiveness}.
However, we argue that the SFT approach has two weaknesses.
First, the fine-tuned LLM is limited to the CSC task.
Second, the fine-tuning procedure requires significant computational resources, even when only a small fraction of parameters are trained.

This work proposes a simple training-free prompt-free framework to leverage LLMs for the CSC task, consisting of two components: an LLM and a distortion model, as shown in Figure~\ref{fig:illustration}.
The key idea is using the LLM as a pure language model in a conventional manner.
The LLM goes through the input sentence from the beginning, and at each inference step, produces a distribution over its vocabulary for deciding the next token, given a partial sentence.
The distortion model ensures that the resulting sentence is faithful to the input sentence, i.e., remaining the same meaning, by capturing pronunciation or shape similarity between the original and replaced characters.

Our contributions are summarized as follows:
\begin{asparaitem}[$\bullet$]
    \item We for the first time propose a simple yet effective training-free prompt-free framework to leverage LLMs for the CSC task, which is totally different from all previous approaches.
    \item Tokens in the output vocabulary of LLMs vary in length, i.e., a token may consist of one or multiple characters. To accommodate this, we propose a length reward, which is very useful and can work well with beam search decoding.
    \item LLM tends to prefer high-frequency tokens, leading to the over-correction issue. We propose a faithfulness reward, which further encourages the model to be faithful successfully.
    \item Experiments on five public datasets demonstrate that our approach significantly improves the performance of LLMs in the CSC task and exhibits remarkable domain generalization capabilities.
\end{asparaitem}

\ifarxiv%
    Our code is available at \url{https://github.com/Jacob-Zhou/simple-csc}.
\fi%

    \section{Our Approach}

Given an input sentence $\boldsymbol{x}={x_1, x_2, \cdots, x_n}$, where $x_i$ denotes a character, a CSC model outputs a sentence of the same length, denoted as $\boldsymbol{y} = {y_1, y_2, \cdots, y_n}$.
The key to the CSC task is how to model the score of the input and output sentence pair, i.e., $\texttt{score}(\boldsymbol{x}, \boldsymbol{y})$.

Under a perspective of probabilistic modeling, $p(\boldsymbol{x}, \boldsymbol{y})$ can be decomposed into two parts:
\begin{equation}
    \begin{aligned}
        p(\boldsymbol{x}, \boldsymbol{y}) & =                    p(\boldsymbol{x}\mid \boldsymbol{y})\,p(\boldsymbol{y})           \\
                                          & = p_{\mathtt{DM}}(\boldsymbol{x}\mid \boldsymbol{y})\,p_{\mathtt{LLM}}(\boldsymbol{y})
        \label{eq:ncm_token}
    \end{aligned}
\end{equation}

The first part corresponds to a distortion model, which captures the relationships between $\boldsymbol{x}$ and $\boldsymbol{y}$. In other words, it interprets how spelling errors transform  $\boldsymbol{y}$ to $\boldsymbol{x}$.
Another important function of the distortion model is to make sure that $\boldsymbol{y}$ represents the same ``meaning'' as $\boldsymbol{x}$, i.e., faithfulness.

The second part corresponds to a large language model, which makes sure that  $\boldsymbol{y}$ is fluent and correct from the language use perspective.

Please note that our use of LLMs is \textbf{prompt-free}.
We \ul{\emph{do not provide CSC-related instructions and examples as the prompt}}.
More importantly, we \ul{\emph{do not give the input sentence to LLMs}}.
We use LLMs as pure traditional language models for evaluating next-token probabilities.

\begin{table}[tp!]
    \centering
    \scalebox{0.9}{
        \begin{tabular}{lrlc}
            \toprule
            \textbf{Type}                  & \multicolumn{2}{c}{\textbf{Example}}                          & \textbf{Proportion}                                                       \\
            \midrule
            \texttt{Identical}             & 机                                                             & (\textit{jī})                                                     & 0.962 \\
            \midrule
            \texttt{Same \rlap{Pinyin}}    & 基                                                             & (\textit{\textcolor{figure_blue}{jī}})                            & 0.023 \\
            \texttt{Similar \rlap{Pinyin}} & 七                                                             & (\textit{\textcolor{figure_orange}{q}\textcolor{figure_blue}{ī}}) & 0.008 \\
            \texttt{Similar \rlap{Shape}}  & \phantom{(\textit{zhǎng})}\llap{\textcolor{figure_orange}{仉}} & (\textit{zhǎng})                                                  & 0.004 \\
            \midrule
            \texttt{Unrelated}             & 能                                                             & (\textit{néng})                                                   & 0.003 \\
            \bottomrule
        \end{tabular}
    }
    \caption{
        Examples of the different distortion types of the corrected token ``机'' (\textit{jī}).
        The distribution of the types is calculated from the development set.
    }
    \label{tab:types}
\end{table}

\subsection{A Minimal Distortion Model}
\label{sec:distortion_model}

Our distortion model adopts character-level factorization:
\begin{equation}
    \log p_{\mathtt{DM}}(\boldsymbol{x}\mid \boldsymbol{y}) =
    \sum_{i} \log p_{\mathtt{DM}}(x_i\mid y_i)
\end{equation}

To further simplify the model, we \textbf{do not} compute distortion probabilities for specific character pairs, i.e., $(c_1, c_2)$. Instead, we first classify $(c_1, c_2)$ into one of five distortion types, denoted as $\texttt{type}(c_1, c_2)$. Then we use the probability of the type as the distortion probability of the character pair:
\begin{equation}
    p_{\mathtt{DM}}(c_1\mid c_2) = p(\texttt{type}(c_1, c_2))
\end{equation}

Table~\ref{tab:types} illustrates the distortion types.
The proportions are obtained from small subsets of popular CSC training data, described later in \S\ref{sec:datasets}, and used directly as distortion probabilities.

Please note that we claim our approach as \textbf{training-free}, since \ul{\textit{the LLMs are used in an off-the-shelf manner}} and \ul{\textit{the distortion model only relies on several frequency values}}, which can be easily counted from a small dataset.

Given $(c_1, c_2)$, we implement a simple rule-based tool to decide the distortion type.
Among the five types, ``\texttt{Similar Pinyin}'' and ``\texttt{Similar Shape}'' are more complex to handle.
More details are given in Appendix \ref{app:detailed_type}.

\subsection{Next-token Probabilities from LLM}
\label{sec:llm_adopting}

Typically, the output vocabulary of an LLM contains both single- and multi-character tokens.
In other words, given a sentence $\boldsymbol{y} = y_1...y_n$, there exists many ways to segment it into a sequence of tokens.
We use $\boldsymbol{t} = t_1...t_m$ to denote a specific token-level segmentation of $\boldsymbol{y}$, i.e., a path for the LLM to generate the character sequence, where $t_j = c_1 \dots c_k$ and $k \ge 1$.
Then, the log probability of $\boldsymbol{y}$ can be decomposed as:
\begin{equation}
    \log p_{\mathtt{LLM}}(\boldsymbol{y})
    = \sum_{j} \log p_{\mathtt{LLM}}(t_j\mid \boldsymbol{t}_{<j})
\end{equation}

After combining the distortion model, the probability of a partial output sentence is:
\begin{equation}
    \begin{aligned}
        \log p(\boldsymbol{x}, \boldsymbol{t}_{\le j}) =\  &
        \log p(\boldsymbol{x}, \boldsymbol{t}_{<j})                                                                 \\
                                                           & + \log p_{\mathtt{LLM}}(t_j\mid \boldsymbol{t}_{<j})   \\
                                                           & + \sum_{r=1}^k \log p_{\mathtt{DM}}(c_r \mid  x_{l+r})
    \end{aligned}
    \label{eq:vanilla_score}
\end{equation}
where $k\!=\!\mathtt{len}(t_j)$ and $l\!=\!\mathtt{len}(\boldsymbol{t}_{<j})$ are the lengths (i.e., character number) of $t_j$ and $\boldsymbol{t}_{<j}$, respectively.

\subsection{Beam Search Decoding}
During inference, the basic operation at step $j$ is to select a token $t_j$ and append it to the current partial sequence $\boldsymbol{t}_{<j}$.
We follow the standard practice, and adopt beam search decoding, that only retains the top-$K$ candidates at each decoding step for computational efficiency.

In particular, one technical detail is closely related with our length reward strategy and thus worthy of further discussion.
As discussed above, most LLMs generate sentences at token-level and one token may contain either a single character or multiple characters.
This implies that the beam search procedure is aligned according to token numbers rather than character positions.
In other words, at any given inference step, candidates in the beam may varies greatly in the number of characters generated so far.
For instance, one candidate contains 5 characters, whereas another candidate contains 8.

\begin{figure}[tb!]
    \newcommand\dc[1]{\textcolor{figure_gray}{#1}}
    \newcommand\wc[1]{\textcolor{figure_red}{#1}}
    \newcommand\cc[1]{\textcolor{figure_blue!90!black}{#1}}
    \newcommand\pruned{\textcolor{black!40}{\ding{55}}}
    \newcommand\finished{\textcolor{figure_blue}{\ding{51}}}
    \captionsetup[subfigure]{skip=3pt}
    \centering
    \subfloat[w/o Length Reward]%
    {
        \label{fig:decode:steps:wx_length}
        \centering
        \scalebox{1.1}{
            \begin{tikzpicture}[
                    font=\scriptsize,
                    label/.style={
                            font=\itshape\tiny,
                            anchor=west,
                            draw=none,
                            inner sep=-0.5pt,
                        },
                    anchor point/.style={
                            draw=none,
                            circle,
                            inner sep=1.5pt,
                        },
                ]
                \graph [
                empty/.style={draw=none},
                successful exploratory/.style={edges={figure_blue, thick}},
                unsuccessful exploratory/.style={edges={figure_gray, semithick, densely dashed}},
                nodes={anchor=base, inner sep=1pt},
                edges={arrows={-Straight Barb[length=0.6mm]}, figure_gray, semithick, out=0, in=180, looseness=0.65},
                grow right sep=1.0cm,
                branch down=0.35,
                multi] {
                {bos/"\texttt{BOS}"} -!- [complete bipartite]
                {x0_0/"要",x0_1/"要求",x0_2/"\dc{姚}",x0_3/"\dc{约}",x0_4/"\dc{药}",x0_5/"\dc{妖}",x0_6/"\dc{腰}",x0_7/"\dc{于}"} -!- [complete bipartite]
                {x1_0/"求",x1_1/"\dc{是}",x1_2/"\wc{师}",x1_3/"\dc{修}",x1_4/"\dc{就}",x1_5/"\dc{实}",x1_6/"\dc{球}",x1_7/"\dc{时}"} -!- [complete bipartite]
                {x2_0/"\wc{公}",x2_1/"\wc{公}",x2_2/"\wc{师}",x2_3/"\dc{是}",x2_4/"\dc{式}",x2_5/"\dc{使}",x2_6/"\dc{实}",x2_7/"\wc{公}"} -!- [complete bipartite]
                {x3_0/"$\cdots$",x3_1/"$\cdots$",x3_2/"$\cdots$",x3_3/"$\cdots$",x3_4/"$\cdots$",x3_5/"$\cdots$",x3_6/"$\cdots$",x3_7/"$\cdots$"} -!- [complete bipartite]

                { [unsuccessful exploratory] bos -> {x0_2,x0_3,x0_4,x0_5,x0_6,x0_7}};
                {bos -> {x0_0,x0_1}};

                {x0_0 -> x1_0};
                { [unsuccessful exploratory] x0_1 -> x1_1};
                { x0_1 -> x1_2};
                { [unsuccessful exploratory] x0_0 -> x1_3};
                { [unsuccessful exploratory] x0_0 -> x1_4};
                { [unsuccessful exploratory] x0_1 -> x1_5};
                { [unsuccessful exploratory] x0_0 -> x1_6};
                { [unsuccessful exploratory] x0_1 -> x1_7};

                {x1_2 -> x2_0};
                { x1_1 -> x2_1};
                { x1_0 -> x2_2};
                { [unsuccessful exploratory] x1_0 -> x2_3};
                { [unsuccessful exploratory] x1_0 -> x2_4};
                { [unsuccessful exploratory] x1_0 -> x2_5};
                { [unsuccessful exploratory] x1_0 -> x2_6};
                { x1_5 -> x2_7};

                { x2_2 -> x3_0};
                { x2_0 -> x3_1};
                { x2_1 -> x3_2};
                { x2_0 -> x3_3};
                { [unsuccessful exploratory] x2_0 -> x3_4};
                { x2_3 -> x3_5};
                { x2_1 -> x3_6};
                { [unsuccessful exploratory] x2_0 -> x3_7};
                };

                \node[label] at (x0_2.east) {\pruned};
                \node[label] at (x0_3.east) {\pruned};
                \node[label] at (x0_4.east) {\pruned};
                \node[label] at (x0_5.east) {\pruned};
                \node[label] at (x0_6.east) {\pruned};
                \node[label] at (x0_7.east) {\pruned};

                \node[label] at (x1_3.east) {\pruned};
                \node[label] at (x1_4.east) {\pruned};
                \node[label] at (x1_6.east) {\pruned};
                \node[label] at (x1_7.east) {\pruned};

                \node[label] at (x2_4.east) {\pruned};
                \node[label] at (x2_5.east) {\pruned};
                \node[label] at (x2_6.east) {\pruned};
                \node[label] at (x2_7.east) {\pruned};
            \end{tikzpicture}
        }
    }\\[5pt]%
    \subfloat[w/ Length Reward]%
    {
        \label{fig:decode:steps:w_length}
        \centering
        \scalebox{1.1}{
            \begin{tikzpicture}[
                    font=\scriptsize,
                    label/.style={
                            font=\itshape\tiny,
                            anchor=west,
                            draw=none,
                            inner sep=-0.5pt,
                        },
                    anchor point/.style={
                            draw=none,
                            circle,
                            inner sep=1.5pt,
                        },
                ]
                \graph [
                empty/.style={draw=none},
                successful exploratory/.style={edges={figure_blue, thick}},
                unsuccessful exploratory/.style={edges={figure_gray, semithick, densely dashed}},
                nodes={anchor=base, inner sep=1pt},
                edges={arrows={-Straight Barb[length=0.6mm]}, figure_gray, semithick, out=0, in=180, looseness=0.65},
                grow right sep=1.1cm,
                branch down=0.35,
                multi] {
                {bos/"\texttt{BOS}"} -!- [complete bipartite]
                {x0_0/"要求",x0_1/"要",x0_2/"\dc{姚}",x0_3/"\dc{约}",x0_4/"\dc{药}",x0_5/"\dc{妖}",x0_6/"\dc{腰}",x0_7/"\dc{于}"} -!- [complete bipartite]
                {x1_0/"\cc{施}\cc{工}单位"[fill=figure_blue!10,rounded corners=2pt,draw=black],x1_1/"\dc{是}",x1_2/"\wc{师}",x1_3/"求",x1_4/"\dc{实}",x1_5/"\dc{时}",x1_6/"\dc{式}",x1_7/"\cc{施}"[fill=figure_blue!10,rounded corners=2pt,draw=black]} -!- [complete bipartite]
                {x2_0/"$\cdots$",x2_1/"$\cdots$",x2_2/"$\cdots$",x2_3/"$\cdots$",x2_4/"$\cdots$",x2_5/"$\cdots$",x2_6/"$\cdots$",x2_7/"$\cdots$"} -!- [complete bipartite]

                { [unsuccessful exploratory] bos -> {x0_2,x0_3,x0_4,x0_5,x0_6,x0_7}};
                {bos -> {x0_0,x0_1}};

                { [successful exploratory] x0_0 -> x1_0};
                { [unsuccessful exploratory] x0_0 -> x1_1};
                { x0_0 -> x1_2};
                { x0_1 -> x1_3};
                { [unsuccessful exploratory] x0_0 -> x1_4};
                { [unsuccessful exploratory] x0_0 -> x1_5};
                { [unsuccessful exploratory] x0_0 -> x1_6};
                { [successful exploratory] x0_0 -> x1_7};

                { x1_0 -> x2_0};
                { x1_2 -> x2_1};
                { x1_1 -> x2_2};
                { x1_3 -> x2_3};
                { x1_3 -> x2_4};
                { [unsuccessful exploratory] x1_1 -> x2_5};
                { x1_4 -> x2_6};
                { x1_7 -> x2_7};
                };

                \node[label] at (x0_2.east) {\pruned};
                \node[label] at (x0_3.east) {\pruned};
                \node[label] at (x0_4.east) {\pruned};
                \node[label] at (x0_5.east) {\pruned};
                \node[label] at (x0_6.east) {\pruned};
                \node[label] at (x0_7.east) {\pruned};

                \node[label] at (x1_5.east) {\pruned};
                \node[label] at (x1_6.east) {\pruned};

            \end{tikzpicture}
        }
    }%
    \caption{
        A real example of the decoding process for the input sentence ``要求\wc{师公}单位对...'' (\textit{Requesting the \wc{master} unit to ...}).
        Here, ``施工'' (\textit{shīgōng}, \textit{construction}) is misspelled as ``师公'' (\textit{shīgōng}).
        Without the length reward, the correct character ``施'' is fail to be select into the beam.
    }
    \label{fig:decode:steps}
\end{figure}
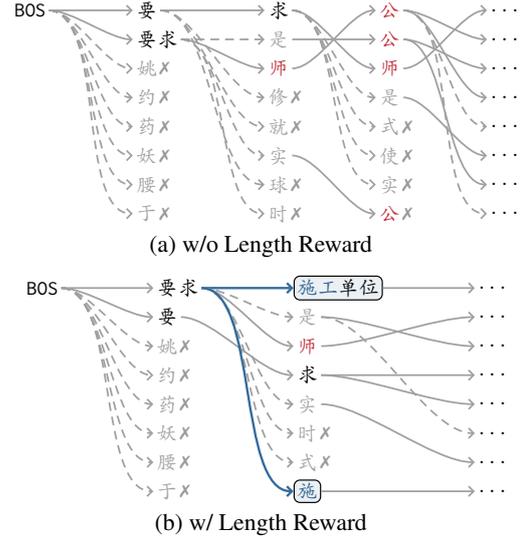

\subsection{Length Reward}
\label{sec:length_reward}
Our preliminary experiments show that the vanilla approach, as described in Equation~\ref{eq:vanilla_score}, produces unsatisfactory results.
Detailed analysis shows that the paths explored in the beam search space are dominated by single-character tokens, as shown in Figure \ref{fig:decode:steps:wx_length}.
As we all know, multi-character tokens are created by merging characters that frequently occur together, capturing the most common patterns in the language.
LLMs are trained for and, in turn, very good at generating multi-character tokens.
Therefore, it is counter-intuitive to deprive such capability from LLMs.

To handle the issue, we design a simple length reward so that the model favors and keeps multi-char tokens during beam search:
\begin{equation}
    \begin{aligned}
        \texttt{score}(\boldsymbol{x}, \boldsymbol{t}_{\le j}) =\  &
        \texttt{score}(\boldsymbol{x}, \boldsymbol{t}_{<j})                                                                                     \\
                                                                   & + \log p_{\mathtt{LLM}}(t_j\mid \boldsymbol{t}_{<j})                       \\
                                                                   & + \sum_{r=1}^k \log p_{\mathtt{DM}}(c_r \mid  x_{l+r})                     \\
                                                                   & + \textcolor{figure_red}{\alpha \times \left( \mathtt{len}(t_i)-1 \right)} \\
    \end{aligned}
\end{equation}
where $\alpha$ is a hyperparameter for balancing the weight of the length reward, considering that the other two components use log probabilities, whereas the length reward uses numbers directly.
Please note that we use $\texttt{score}(\cdot)$ instead of $p(\cdot)$, since the values are no longer probabilities.

As shown in Figure \ref{fig:decode:steps:w_length}, thanks to the length reward, the correct token ``施工单位'' (\textit{construction unit}) is now ranked within the top-$K$ candidates.
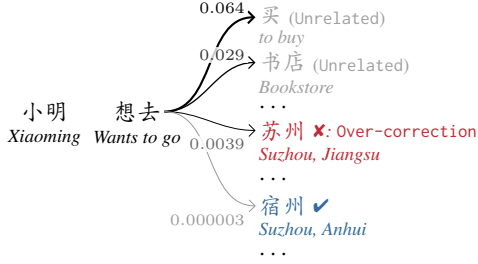
\begin{figure}[tb!]
    \captionsetup[subfigure]{skip=3pt}
    \centering
    \scalebox{1.0}{
        \begin{tikzpicture}[
                font=\footnotesize,
                label/.style={
                        font=\itshape\scriptsize,
                        draw=none,
                        fill=white,
                        fill opacity=0.8,
                        text opacity=1.0,
                        inner sep=1pt,
                    },
                anchor point/.style={
                        draw=none,
                        circle,
                        inner sep=1.5pt,
                    },
            ]
            \graph [
            empty/.style={draw=none},
            nodes={anchor=base, inner sep=1pt},
            edges={arrows = {-Straight Barb[length=0.8mm]}},
            grow right sep=0.5cm,
            level 2/.style={grow right sep=1.2cm},
            branch down=0.5,
            multi] {
            x1/"小明" [yshift=-1.25cm] -!- x2/"想去" [yshift=-1.25cm] ->[out=0, in=180] {
            x3_0/"\textcolor{figure_gray}{买 \scriptsize{(\texttt{Unrelated})}}"[target edge style={line width=0.8, draw=black!100}],
            x3_1/"\textcolor{figure_gray}{书店 \scriptsize{(\texttt{Unrelated})}}"[target edge style={line width=0.5, draw=black!100}, yshift=-0.1cm],
            x3_2/"$\cdots$" [target edge style={line width=0.5, draw=none},text=black, yshift=-0.2cm],
            x3_3/"\textcolor{figure_red}{苏州}"[target edge style={line width=0.3, draw=black!100}],
            x3_4/"$\cdots$"[target edge style={draw=none}, yshift=-0.15cm],
            x3_5/"\textcolor{figure_blue}{宿州}"[target edge style={line width=0.2, draw=black!40}],
            x3_6/"$\cdots$"[target edge style={draw=none}, yshift=-0.15cm],
            };
            };
            \node[label, below=0.0cm of x1, inner sep=1pt] {Xiaoming\strut};
            \node[label, below=0.0cm of x2, inner sep=1pt] {Wants to go\strut};
            \node[label, anchor=north west, inner sep=1pt] at ($(x3_0.south west) + (0, 0.05cm)$) {\textcolor{figure_gray}{to buy}};
            \node[label, anchor=east, inner sep=1pt, font=\tiny, color=black!100, fill=white] at ($(x3_0.south west) - (0.1, -0.3)$) {$0.064$};
            \node[label, anchor=north west, inner sep=1pt] at ($(x3_1.south west)$) {\textcolor{figure_gray}{Bookstore}};
            \node[label, anchor=east, inner sep=1pt, font=\tiny, color=black!80, fill=white] at ($(x3_1.south west) - (0.1, -0.3)$) {$0.029$};
            \node[label, anchor=north west, inner sep=1pt] at ($(x3_3.south west)$) {\textcolor{figure_red}{Suzhou, Jiangsu}};
            \node[label, anchor=east, inner sep=1pt, font=\tiny, color=black!60, fill=white] at ($(x3_3.south west) - (0.1, 0)$) {$0.0039$};
            \node[label, anchor=north west, inner sep=1pt] at ($(x3_5.south west)$) {\textcolor{figure_blue}{Suzhou, Anhui}};
            \node[label, anchor=east, inner sep=1pt, font=\tiny, color=black!40, fill=white] at ($(x3_5.south west) - (0.1, 0)$) {$0.000003$};
            \node[label, anchor=west, inner sep=1pt] at ($(x3_3.east)$) {\textcolor{figure_red}{\textrm{\ding{56}:}\ \texttt{Over-correction}}};
            \node[label, anchor=west, inner sep=1pt] at ($(x3_5.east)$) {\textcolor{figure_blue}{\ding{52}}};
        \end{tikzpicture}
    }
    \caption{
        A real example of the probabilities for the next token, given the partial sequence ``小明想去'' from the sentence ``小明想去\textcolor{figure_blue}{宿州}'' (\textit{Xiaoming wants to go to \textcolor{figure_blue}{Suzhou, Anhui}}).
    }
    \label{fig:faithfulness_reward}
\end{figure}

\subsection{Faithfulness Reward}
\label{sec:confidence_adjustment}
Under our prompt-free use, the LLM component is unaware of the input sentence, and only focuses on the fluency and correctness of the output sentence from the language use perspective.

We observe that our approach, even with the length reward, tends to over-correct the input sentence, i.e., changing its original meaning.
Figure~\ref{fig:faithfulness_reward} gives an example.
Given the partial output sentence, i.e., ``小明想去'' (\textit{Xiaoming wants to go to}), the LLM component gives a probability of $0.0039$ to ``苏州'' (\textit{sūzhōu}), which is a very famous city in Jiangsu Province.
In contrast, it gives a much lower probability of $\num{3d-6}$ to the original input token, i.e., ``宿州'' (\textit{sùzhōu}), which is a less famous city in Anhui Province.
The distortion model fails to remedy such great gap.
As the result, our approach adopts the ``correction''.
However, under such circumstances, it is better to reserve the original tokens.

To mitigate this issue, we introduce a faithfulness reward:
\begin{equation}
    \begin{aligned}
         & \texttt{score}(\boldsymbol{x}, \boldsymbol{t}_{\le j}) =
        \texttt{score}(\boldsymbol{x}, \boldsymbol{t}_{<j})                                                                                         \\
         & \ + \log p_{\mathtt{LLM}}(t_j\mid \boldsymbol{t}_{<j})                                                                                   \\
         & \ + \textcolor{figure_red}{(1+H_{\mathtt{LLM}(\cdot)})}\times\left(\begin{array}{c}
                                                                                      \sum_{r=1}^k \log p_{\mathtt{DM}}(c_r \mid  x_{l+r}) \\
                                                                                      +                                                    \\
                                                                                      \alpha\times\left( \mathtt{len}(t_i)-1 \right)\end{array}\right)
    \end{aligned}
\end{equation}
where $H_{\mathtt{LLM}(\cdot)}$ denote the entropy of next-token probabilities.\footnote{Since LLMs have different output vocabularies $\mathcal{V}$, we divide the entropy by $\log |\mathcal{V}|$, which can be understood as the maximum entropy, and the value will fall into $[0, 1]$. }
If the entropy is high, meaning that the LLM is uncertain about the next token, the distortion model, along with the length reward, will play a more important role in deciding the next token. From Table \ref{tab:types}, we can see that the ``\texttt{Identical}'' type has a much higher probability than others. That is, the distortion model always favors the original input tokens.

    \begin{table*}[tb!]
    \setlength{\tabcolsep}{3.85pt}
    \renewcommand{\arraystretch}{0.95}
    \centering
    \scalebox{0.92}{
        \begin{NiceTabular}{lccc>{\columncolor{figure_light_red!6}}c|cc>{\columncolor{figure_light_red!6}}c|cc>{\columncolor{figure_light_red!6}}c|cc>{\columncolor{figure_light_red!6}}c|cc>{\columncolor{figure_light_red!6}}c}
            \toprule
            \rowcolor[gray]{1.0}
            \Block[l]{2-2}{\textbf{System}}                                                                                     &              & \Block[c]{1-3}{\textbf{rSighans}} &               &                 & \Block[c]{1-3}{\textbf{CSCD-NS}} &               &               & \Block[c]{1-3}{\textbf{MCSCSet}} &               &                 & \Block[c]{1-3}{\textbf{ECSpell}} &               &                 & \Block[c]{1-3}{\textbf{Lemon}} &               &                 \\
            \rowcolor[gray]{1.0}
                                                                                                                                &              & S-F\bgood                         & C-F\bgood     & FPR\sgood       & S-F\bgood                        & C-F\bgood     & FPR\sgood     & S-F\bgood                        & C-F\bgood     & FPR\sgood       & S-F\bgood                        & C-F\bgood     & FPR\sgood       & S-F\bgood                      & C-F\bgood     & FPR\sgood       \\
            \midrule
            \rowcolor[gray]{.95}
            \Block[l]{1-16}{\texttt{Domain-Specific\;SOTAs} (\textit{Trained on in-domain gold-standard data of each dataset})} &              &                                   &               &                 &                                  &               &               &                                  &               &                 &                                  &               &                 &                                &               &                 \\
            \Block[l]{1-2}{\texttt{ReaLiSe}$^\dagger$}                                                                          &              & 69.3                              & 80.7          & 10.1            & \ood{41.4}                       & \ood{44.2}    & \ood{27.6}    & \ood{17.8}                       & \ood{27.6}    & \ood{12.0}      & \ood{34.9}                       & \ood{45.4}    & \ood{13.7}      & \ood{28.2}                     & \ood{31.6}    & \ood{19.1}      \\
            \Block[l]{1-2}{\citet{hu-etal-2024-cscd}}                                                                           &              & --                                & --            & --              & 74.4                             & 76.6          & --            & --                               & --            & --              & --                               & --            & --              & --                             & --            & --              \\
            \Block[l]{1-2}{\citet{jiang-etal-2022-mcscset}}                                                                     &              & --                                & --            & --              & --                               & --            & --            & 80.9                             & --            & --              & --                               & --            & --              & --                             & --            & --              \\
            \Block[l]{1-2}{\citet{liu-etal-2023-chinese}}                                                                       &              & --                                & --            & --              & --                               & --            & --            & --                               & --            & --              & 85.7                             & --            & \wz5.4          & --                             & --            & --              \\
            \midrule
            \rowcolor[gray]{.95}
            \Block[l]{1-16}{\texttt{Domain-General\;SOTAs} (\textit{Trained on about 34M synthetic CSC data})}                  &              &                                   &               &                 &                                  &               &               &                                  &               &                 &                                  &               &                 &                                &               &                 \\
            \Block[l]{1-2}{\texttt{Finetuned\;BERT}}                                                                            &              & 47.5                              & \textbf{57.5} & 16.9            & \textbf{52.0}                    & \textbf{53.9} & \textbf{25.7} & 35.3                             & 48.5          & \wz7.5          & 57.1                             & 64.9          & \textbf{\wz6.4} & 48.0                           & 49.3          & 13.1            \\
            \Block[l]{1-2}{\texttt{Softmasked\;BERT}}                                                                           &              & \textbf{47.7}                     & 57.4          & 15.1            & 51.0                             & 53.4          & 28.5          & 35.3                             & 48.5          & \wz8.1          & 57.6                             & 66.2          & \wz7.6          & 47.2                           & 48.8          & 13.1            \\
            \Block[l]{1-2}{\texttt{ReLM}}                                                                                       &              & 47.3                              & 56.9          & \textbf{\wz9.6} & 49.5                             & 51.6          & 29.3          & \textbf{37.8}                    & \textbf{50.2} & \textbf{\wz6.8} & \textbf{59.3}                    & \textbf{68.4} & \wz8.6          & \textbf{50.2}                  & \textbf{51.3} & \textbf{11.8}   \\
            \midrule
            \rowcolor[gray]{.95}
            \Block[l]{1-16}{\texttt{LLMs} (\textit{without CSC-specific training})}                                             &              &                                   &               &                 &                                  &               &               &                                  &               &                 &                                  &               &                 &                                &               &                 \\
            \Block[c]{3-1}{\texttt{Baichuan2}                                                                                                                                                                                                                                                                                                                                                                                                                                                        \\[-4pt]\texttt{\footnotesize{(13B)}}}
                                                                                                                                & \texttt{ZSP} & 19.0                              & 18.4          & 49.1            & 22.6                             & 14.5          & 35.3          & 13.6                             & \wz8.0        & 77.5            & 34.5                             & 22.3          & 30.3            & 17.5                           & \wz9.8        & 40.9            \\
                                                                                                                                & \texttt{FSP} & 31.8                              & 38.5          & 21.4            & 35.7                             & 32.7          & \textbf{10.5} & 42.6                             & 47.1          & \wz4.4          & 56.8                             & 53.1          & \wz5.8          & 35.1                           & 25.2          & \wz9.5          \\
                                                                                                                                & \texttt{OUR} & \textbf{59.1}                     & \textbf{70.9} & \textbf{10.4}   & \textbf{63.2}                    & \textbf{66.2} & 16.5          & \textbf{66.0}                    & \textbf{76.9} & \textbf{\wz1.7} & \textbf{84.5}                    & \textbf{89.8} & \textbf{\wz4.9} & \textbf{53.2}                  & \textbf{56.2} & \textbf{\wz9.1} \\
            \hdashedline
            \Block[c]{3-1}{\texttt{Qwen1.5}                                                                                                                                                                                                                                                                                                                                                                                                                                                          \\[-4pt]\texttt{\footnotesize{(14B)}}}
                                                                                                                                & \texttt{ZSP} & 29.0                              & 31.4          & 41.1            & 34.3                             & 31.3          & 24.5          & 40.2                             & 45.4          & \wz3.8          & 50.9                             & 49.0          & 14.4            & 31.8                           & 26.8          & 16.1            \\
                                                                                                                                & \texttt{FSP} & 34.3                              & 37.9          & 26.2            & 42.9                             & 38.7          & \textbf{10.4} & 40.5                             & 44.3          & \textbf{\wz3.1} & 59.0                             & 58.2          & \textbf{\wz5.9} & 37.2                           & 30.2          & \textbf{\wz9.9} \\
                                                                                                                                & \texttt{OUR} & \textbf{54.4}                     & \textbf{68.0} & \textbf{17.2}   & \textbf{52.6}                    & \textbf{57.7} & 25.8          & \textbf{61.1}                    & \textbf{72.6} & \textbf{\wz3.1} & \textbf{81.6}                    & \textbf{88.2} & \wz6.5          & \textbf{46.3}                  & \textbf{50.8} & 14.1            \\
            \hdashedline
            \Block[c]{3-1}{\texttt{InternLM2}                                                                                                                                                                                                                                                                                                                                                                                                                                                        \\[-4pt]\texttt{\footnotesize{(20B)}}}
                                                                                                                                & \texttt{ZSP} & 31.0                              & 30.4          & 57.3            & 34.9                             & 29.2          & 40.6          & 19.0                             & 12.5          & 80.5            & 45.2                             & 37.5          & 31.6            & 32.8                           & 26.5          & 27.8            \\
                                                                                                                                & \texttt{FSP} & 35.2                              & 38.8          & 31.7            & 39.4                             & 35.1          & 22.4          & 33.6                             & 32.6          & 20.4            & 54.3                             & 49.8          & 15.7            & 35.9                           & 28.9          & 17.3            \\
                                                                                                                                & \texttt{OUR} & \textbf{57.1}                     & \textbf{70.0} & \textbf{12.6}   & \textbf{60.7}                    & \textbf{64.1} & \textbf{19.7} & \textbf{63.2}                    & \textbf{72.9} & \textbf{\wz2.6} & \textbf{82.4}                    & \textbf{88.8} & \textbf{\wz5.1} & \textbf{49.8}                  & \textbf{53.7} & \textbf{10.7}   \\
            \bottomrule
        \end{NiceTabular}
    }
    \caption{
        Main Results.
        $\dagger$: We reran the released code of \texttt{ReaLiSe} \cite{xu-etal-2021-read}, along with their released models, to obtain the results.
        \texttt{ReaLiSe}, was trained on the in-domain, gold-standard data of the Sighans dataset and represents a SOTA model for it.
        The numbers in \ood{gray} represent the out-of-domain results for \texttt{ReaLiSe}.
        Detailed results of each sub-domain are provided in Appendix~\ref{app:subdomain_results}.
    }
    \label{tab:main:results}
\end{table*}

\section{Experimental Setup}
\subsection{Datasets.}
\label{sec:datasets}

\paragraph{Pseudo development set}
Since there is no publicly available, manually labeled, domain-general development set for CSC, we have chosen to split a small portion of the existing synthetic training data for hyperparameter tuning, naming it \textbf{Pseudo-Dev}.
Specifically, we use 1,000 sentences each from the synthetic training data of \citet{hu-etal-2024-cscd} and \citet{wang-etal-2018-hybrid} as our development set.

\paragraph{Real-world test sets}
We perform experiments across five distinct CSC datasets: \textbf{Sighans} \cite{wu-etal-2013-chinese,yu-etal-2014-overview,tseng-etal-2015-introduction}, \textbf{CSCD-NS} \cite{hu-etal-2024-cscd}, \textbf{MCSCSet} \cite{jiang-etal-2022-mcscset}, \textbf{ECSpell} \cite{lv-etal-2023-ecspell}, and \textbf{Lemon} \cite{wu-etal-2023-rethinking}, covering a broad spectrum of domains and genres.
The details and statistics of these datasets can be found in Appendix~\ref{app:test_sets}.
For Sighans, we utilize the revised versions released by \citet{yang-etal-2023-chinese}, which have been manually verified and corrected for errors of the original datasets, and name them as \textbf{rSighans} for clarity.

\paragraph{Selected datasets for analyses}
\label{para:selected_datasets}
Given the absence of a domain-general development set for CSC and the potential limitations of the \textbf{Pseudo-Dev} set in representing real-world data, we conduct in-depth analyses on three distinct datasets to cover a broad spectrum of language use.
These include errors made by Chinese learners (\textbf{rSighan}~\textit{15}), colloquial and diverse text from novels (\textbf{Lemon}~\textit{Nov}), and formal and standard text from official documents (\textbf{ECSpell}~\textit{Odw}).

\subsection{Evaluation Metrics.}
We follow the convention to use the \textbf{sentence-level correction} $F_1$ (\textbf{S-F}) score as the main evaluation metric.
Besides, we also report \textbf{character-level correction} $F_1$ (\textbf{C-F}) and \textbf{sentence-level false positive rate}~(\textbf{FPR}) to provide a more complete view of the model performance.
Details of the evaluation metrics can be found in Appendix~\ref{app:evaluation_metrics}.

\subsection{Baselines}
We compare our approach against prompt-based method under two settings: zero-shot prompting (\textbf{ZSP}) and few-shot prompting (\textbf{FSP}).
For few-shot settings, we select 10 examples from the \textbf{Pseudo-Dev}.
The details of the prompts can be found in Appendix~\ref{app:prompt_examples}, and the example selection strategy is described in Appendix~\ref{app:few_shot_examples}.
During inference, we adopt the greedy decoding strategy\rlap{.}\footnote{We observe that the improvement of beam search is marginal and sometimes even detrimental.}

To provide a more comprehensive comparison, we also present results from state-of-the-art domain-general CSC models trained on 34 million pairs of synthetic CSC data for reference.
These models include \textbf{Finetuned BERT} \cite{devlin-etal-2019-bert}, \textbf{Softmasked BERT} \cite{zhang-etal-2020-spelling}, and \textbf{ReLM} \cite{liu-etal-2023-chinese}\rlap{.}\footnote{The results of these models were obtained by running the released code along with the corresponding checkpoints provided at \url{https://github.com/gingasan/lemon.git}.}

Additionally, for datasets that have in-domain manually annotated data, we report results from models specifically trained on it, serving as another reference point.

\subsection{Selection of LLMs}
We conduct experiments on three open-source LLMs: \texttt{Baichuan2} \cite{yang-etal-2023-baichuan}, \texttt{Qwen1.5} \cite{bai-etal-2023-qwen}, and \texttt{InternLM2} \cite{cai-etal-2024-internlm2}.
For the main results, we select models with parameter sizes ranging from 10B to 20B to ensure that the LLMs have sufficient zero-shot and few-shot capabilities for meaningful comparisons.
Additionally, we report the ZSP and FSP results of the widely recognized best-performing LLM family, GPT, including \texttt{GPT-3.5} and \texttt{GPT-4}.

To simplify the analysis, we select the \texttt{Bai}-\texttt{chuan2} \texttt{7B} as a representative model to investigate the impact of components in our approach.

\subsection{Hyperparameters of Our Approach}
We use the ``\texttt{Base}'' version of each LLM family.
The distortion probabilities of distortion model were derived from the statistics of the Pseudo-Dev dataset.
We tuned $\alpha$ on \texttt{Baichuan2\;7B} using the Pseudo-Dev dataset.
Eventually, $\alpha$ was set to 2.5 for all experiments.
During inference, we adopt beam search with a beam size of 8.

    \begin{table}[tb!]
    \centering
    \renewcommand{\arraystretch}{0.90}
    \scalebox{0.9}{
        \setlength{\tabcolsep}{2.5pt}%
        \begin{NiceTabular}{l@{\;}c>{\columncolor{figure_light_blue!40}}ccc;>{\columncolor{figure_light_blue!40}}ccc;>{\columncolor{figure_light_red!6}}c}
            \toprule
            \rowcolor[gray]{1.0}
            \Block[l]{1-2}{\textbf{System}}                                   &                                  & \textbf{S-F}\bgood & S-P\bgood     & S-R\bgood     & \textbf{C-F}\bgood & C-P\bgood     & C-R\bgood     & \textbf{FPR}\sgood \\
            \midrule
            \rowcolor[gray]{.95}
            \Block[l]{1-2}{\textbf{rSighan} \textit{15}}                      &                                  &                    &               &               &                    &               &               &                    \\
            \midrule
            \texttt{ReLM}                                                     &                                  & 55.5               & 61.1          & 50.8          & 61.0               & 78.5          & 49.9          & \wz9.5             \\
            \midrule
            \Block[l]{2-1}{\texttt{GPT3.5}}                                   & \texttt{ZSP}                     & 42.0               & 41.7          & 42.3          & 47.8               & 42.5          & 54.6          & 25.8               \\
                                                                              & \texttt{FSP}                     & 41.7               & 42.0          & 41.4          & 48.4               & 44.5          & 53.2          & 23.4               \\
            \hdashedline
            \Block[l]{2-1}{\texttt{GPT4}}                                     & \texttt{ZSP}                     & 43.5               & 38.1          & 50.8          & 49.9               & 40.2          & 66.0          & 47.5               \\
                                                                              & \texttt{FSP}                     & 48.7               & 44.2          & 54.4          & 52.9               & 44.0          & \textbf{66.3} & 38.8               \\
            \midrule
            \texttt{BC2\ \;13B}                                               & \hspace{-0.5em}┌┐\hspace{-0.5em} & 59.6               & 66.5          & 54.0          & 67.3               & 78.3          & 59.0          & \textbf{\wz8.3}    \\
            \texttt{Q1.5\;14B}                                                & \texttt{OUR}                     & 57.6               & 62.5          & 53.4          & 66.0               & 74.1          & 59.4          & 10.2               \\
            \texttt{IL2\ \;20B}                                               & \hspace{-0.5em}└┘\hspace{-0.5em} & \textbf{60.5}      & \textbf{67.2} & \textbf{55.0} & \textbf{67.8}      & \textbf{78.7} & 59.6          & \textbf{\wz8.3}    \\
            \midrule
            \midrule
            \rowcolor[gray]{.95}
            \Block[l]{1-2}{\textbf{Lemon} \textit{Nov \footnotesize{(1000)}}} &                                  &                    &               &               &                    &               &               &                    \\
            \midrule
            \texttt{ReLM}                                                     &                                  & 36.4               & 46.7          & 29.8          & 36.0               & 49.2          & 28.3          & 14.3               \\
            \midrule
            \Block[l]{2-1}{\texttt{GPT3.5}}                                   & \texttt{ZSP}                     & 19.1               & 20.8          & 17.7          & 19.8               & 17.9          & 22.3          & 29.2               \\
                                                                              & \texttt{FSP}                     & 25.5               & 31.4          & 21.4          & 24.9               & 27.2          & 23.0          & 19.6               \\
            \hdashedline
            \Block[l]{2-1}{\texttt{GPT4}}                                     & \texttt{ZSP}                     & 30.6               & 28.4          & 33.1          & 33.4               & 26.9          & 44.1          & 33.5               \\
                                                                              & \texttt{FSP}                     & 42.7               & 41.4          & \textbf{44.0} & 43.1               & 38.9          & \textbf{48.3} & 27.4               \\
            \midrule
            \texttt{BC2\ \;13B}                                               & \hspace{-0.5em}┌┐\hspace{-0.5em} & \textbf{45.3}      & \textbf{53.7} & 39.1          & \textbf{49.1}      & \textbf{57.0} & 43.2          & \textbf{13.1}      \\
            \texttt{Q1.5\;14B}                                                & \texttt{OUR}                     & 38.2               & 41.7          & 35.3          & 43.7               & 44.5          & 43.0          & 21.8               \\
            \texttt{IL2\ \;20B}                                               & \hspace{-0.5em}└┘\hspace{-0.5em} & 42.8               & 49.9          & 37.5          & 46.4               & 52.8          & 41.4          & 15.3               \\
            \midrule
            \midrule
            \rowcolor[gray]{.95}
            \Block[l]{1-2}{\textbf{ECSpell} \textit{Odw}}                     &                                  &                    &               &               &                    &               &               &                    \\
            \midrule
            \texttt{ReLM}                                                     &                                  & 66.5               & 67.5          & 65.6          & 73.0               & 86.4          & 63.1          & \wz7.1             \\
            \midrule
            \Block[l]{2-1}{\texttt{GPT3.5}}                                   & \texttt{ZSP}                     & 58.2               & 62.5          & 54.5          & 61.0               & 62.7          & 59.3          & \wz4.6             \\
                                                                              & \texttt{FSP}                     & 59.3               & 64.1          & 55.2          & 60.7               & 62.4          & 59.0          & \wz2.4             \\
            \hdashedline
            \Block[l]{2-1}{\texttt{GPT4}}                                     & \texttt{ZSP}                     & 73.1               & 73.0          & 73.3          & 77.3               & 75.5          & 79.2          & \wz5.0             \\
                                                                              & \texttt{FSP}                     & 73.2               & 73.5          & 72.9          & 78.5               & 78.3          & 78.7          & \wz5.0             \\
            \midrule
            \texttt{BC2\ \;13B}                                               & \hspace{-0.5em}┌┐\hspace{-0.5em} & \textbf{92.0}      & \textbf{94.4} & \textbf{89.7} & \textbf{93.8}      & 95.6          & \textbf{92.1} & \textbf{\wz0.4}    \\
            \texttt{Q1.5\;14B}                                                & \texttt{OUR}                     & 87.4               & 88.6          & 86.3          & 91.6               & 91.8          & 91.3          & \wz2.9             \\
            \texttt{IL2\ \;20B}                                               & \hspace{-0.5em}└┘\hspace{-0.5em} & 91.1               & 92.9          & 89.3          & \textbf{93.8}      & \textbf{95.9} & 91.8          & \textbf{\wz0.4}    \\
            \bottomrule
        \end{NiceTabular}
    }
    \caption{
        The comparison to GPT family on the rSighan 15, Lemon Nov, and ECSpell Odw datasets.
        The version of \texttt{GPT3.5} is `\texttt{gpt-3.5-turbo-0125}', \texttt{GPT4} is `\texttt{gpt-4-0613}'.
        \texttt{BC2} is short for \texttt{Baichuan2},
        \texttt{Q1.5} for \texttt{Qwen1.5}, and
        \texttt{IL2} for \texttt{InternLM2}.
    }
    \label{tab:analysis:gpt}
\end{table}

\section{Main Results}
We present the main results in Table~\ref{tab:main:results} and the comparison to the GPT family in Table~\ref{tab:analysis:gpt}.
Conducting a comprehensive evaluation of the GPT family is expensive, so we limit the comparison to a small-scale study, focusing on the three datasets mentioned in Section~\ref{para:selected_datasets}.\!\footnote{The original Lemon-Nov dataset includes 6,000 sentences, which is excessively large for our scope. Therefore, we selected the first 1,000 sentences for this comparison.}
Moreover, several qualitative examples are provided in Appendix~\ref{app:qualitative_examples} to illustrate the performance of our approach.

After applying our approach, all three LLM families outperforms their prompt-based counterparts on all five datasets by a large margin.

Compared to the recent state-of-the-art domain-general CSC models, which are trained on 34M synthetic CSC data, our approach also achieves competitive or even superior performance on most datasets, especially on the {MCSCSet} and {ECSpell} datasets.
The results indicate that our approach has a better generalization across different domains and genres than the current domain-general SOTAs.
However, our approach still largely lags behind the domain-specific SOTAs trained on the gold-standard labeled data of each dataset.

\label{app:compare_gpt}

Compared to the GPT family, our approach consistently outperforms \texttt{GPT3.5} on all three datasets, and achieves better performance than \texttt{GPT4} in most cases.
However, our approach may exhibit a lower C-R compared to \texttt{GPT4}, indicating that we might miss some errors that \texttt{GPT4} can correct.

    \section{Discussion}

\begin{table}[tb!]
    \setlength{\tabcolsep}{2.0pt}
    \renewcommand{\arraystretch}{0.9}
    \centering
    \scalebox{0.85}{
        \begin{NiceTabular}{lrcc>{\columncolor{figure_light_red!6}}c|cc>{\columncolor{figure_light_red!6}}c|cc>{\columncolor{figure_light_red!6}}c}
            \toprule
            \rowcolor[gray]{1.0}
            \Block[l]{2-2}{\textbf{System}} &                           & \Block[c]{1-3}{\textbf{rSighan} \textit{15}} &               &                 & \Block[c]{1-3}{\textbf{Lemon} \textit{Nov}} &               &               & \Block[c]{1-3}{\textbf{ECSpell} \textit{Odw}} &               &              \\
            \rowcolor[gray]{1.0}
                                            &                           & S-F\bgood                                    & C-F\bgood     & FPR\sgood       & S-F\bgood                                   & C-F\bgood     & FPR\sgood     & S-F\bgood                                     & C-F\bgood     & FPR\sgood    \\
            \midrule
            \Block[l]{2-1}{\texttt{BC2}}
                                            & \uline{\texttt{7B}}       & \textbf{59.8}                                & \textbf{68.2} & \textbf{\wz8.0} & 43.2                                        & 47.7          & 13.6          & 89.7                                          & 93.0          & 1.3          \\
                                            & \texttt{13B}              & 59.6                                         & 67.3          & \wz8.3          & \textbf{43.5}                               & \textbf{47.9} & \textbf{13.0} & \textbf{92.0}                                 & \textbf{93.8} & \textbf{0.4} \\
            \midrule
            \Block[l]{6-1}{\texttt{Q1\!.\!5}}
                                            & \texttt{0\!.\!5B}         & 56.3                                         & 63.5          & 10.0            & 33.2                                        & 40.2          & 22.2          & 84.7                                          & 89.9          & 3.8          \\
                                            & \uline{\texttt{1\!.\!8B}} & 58.3                                         & 65.3          & 10.3            & 35.6                                        & 42.3          & 19.9          & \textbf{90.3}                                 & \textbf{92.8} & \textbf{1.7} \\
                                            & \texttt{4B}               & 58.4                                         & 66.8          & 10.0            & 35.9                                        & 42.3          & 21.1          & 88.4                                          & 91.1          & 3.4          \\
                                            & \uline{\texttt{7B}}       & \textbf{59.4}                                & \textbf{67.0} & \textbf{\wz8.5} & \textbf{39.0}                               & \textbf{44.7} & \textbf{19.0} & 87.1                                          & 91.4          & 3.4          \\
                                            & \texttt{14B}              & 57.6                                         & 66.0          & 10.2            & 36.4                                        & 42.6          & 21.2          & 87.4                                          & 91.6          & 2.9          \\
                                            & \texttt{32B}              & 57.2                                         & 65.8          & 10.0            & 36.6                                        & 42.2          & 19.4          & 88.2                                          & 91.9          & 2.9          \\
            \midrule
            \Block[l]{3-1}{\texttt{IL2}}
                                            & \uline{\texttt{1\!.\!8B}} & 55.3                                         & 64.0          & 12.2            & 33.2                                        & 40.1          & 22.6          & 88.3                                          & 91.0          & 2.1          \\
                                            & \uline{\texttt{7B}}       & 58.1                                         & 65.5          & 10.2            & 38.8                                        & 44.2          & 18.0          & 89.3                                          & 92.0          & 2.1          \\
                                            & \texttt{20B}              & \textbf{60.5}                                & \textbf{67.8} & \textbf{\wz8.3} & \textbf{40.5}                               & \textbf{45.3} & \textbf{15.1} & \textbf{91.1}                                 & \textbf{93.8} & \textbf{0.4} \\
            \bottomrule
        \end{NiceTabular}
    }
    \caption{
        Ablation results of model size.
    }
    \label{tab:ablation:size}
\end{table}

\subsection{Impact of the Size of the LLM}
\label{sec:ablation:size}
First, we investigate the impact of the LLM size on the performance of our approach.

As shown in Table~\ref{tab:ablation:size}, in general, larger LLMs tend to perform better than smaller ones within the same model family.
However, the \texttt{Qwen1.5} model family is an exception: the performance improvement becomes marginal when the model size exceeds 1.8B parameters and even decreases when the model size reaches 7B.

When comparing the performance of models of the same size across different model families, we find that the \texttt{Baichuan2} family generally outperforms the other two model families.

\subsection{Effectiveness of the Distortion Model}
\begin{table}[tb!]
    \centering
    \scalebox{0.85}{
        \setlength{\tabcolsep}{2.6pt}%
        \renewcommand{\arraystretch}{0.92}
        \begin{NiceTabular}{l>{\columncolor{figure_light_blue!40}}ccc;>{\columncolor{figure_light_blue!40}}ccc;>{\columncolor{figure_light_red!6}}c}
            \toprule
            \rowcolor[gray]{1.0}
            \Block[l]{1-1}{\textbf{System}}               & \textbf{S-F}\bgood & S-P\bgood  & S-R\bgood  & \textbf{C-F}\bgood & C-P\bgood  & C-R\bgood  & \textbf{FPR}\sgood \\
            \midrule
            \rowcolor[gray]{.95}
            \Block[l]{1-8}{\textbf{rSighan} \textit{15}}  &                    &            &            &                    &            &            &                    \\
            \midrule
            \texttt{CTG}                                  & \wz\wm6.7          & \wz\wm5.3  & \wz\wm9.1  & \wz\wm7.7          & \wz\wm4.2  & \wm47.7    & \wm90.0            \\
            \midrule
            \texttt{OUR}                                  & \wm59.8            & \wm66.0    & \wm54.7    & \wm68.2            & \wm77.8    & \wm60.6    & \wm\wz8.0          \\
            \hdashedline
            \ \ \texttt{-\,DT}                            & \wz\ewm7.7         & \ewm12.6   & \wz\ewm3.9 & \wz\ewm7.1         & \ewm15.7   & \wz\ewm0.3 & \wz\ewp9.4         \\
            \ \ \texttt{-\,DT}$^\dagger$                  & \ewm12.3           & \ewm18.2   & \wz\ewm7.5 & \wz\ewm9.8         & \ewm20.5   & \wz\ewm1.2 & \ewp11.1           \\
            \midrule
            \midrule
            \rowcolor[gray]{.95}
            \Block[l]{1-8}{\textbf{Lemon} \textit{Nov}}   &                    &            &            &                    &            &            &                    \\
            \midrule
            \texttt{CTG}                                  & \wz\wm0.7          & \wz\wm0.5  & \wz\wm1.1  & \wz\wm1.4          & \wz\wm0.7  & \wm22.5    & \wm96.2            \\
            \midrule
            \texttt{OUR}                                  & \wm43.2            & \wm52.2    & \wm36.9    & \wm47.7            & \wm55.5    & \wm41.9    & \wm13.6            \\
            \hdashedline
            \ \ \texttt{-\,DT}                            & \ewm12.3           & \ewm20.5   & \wz\ewm6.8 & \ewm10.0           & \ewm20.8   & \wz\ewm0.7 & \ewp13.9           \\
            \ \ \texttt{-\,DT}$^\dagger$                  & \ewm11.7           & \ewm20.5   & \wz\ewm5.5 & \wz\ewm9.7         & \ewm21.8   & \wz\ewm1.6 & \ewp14.7           \\
            \midrule
            \midrule
            \rowcolor[gray]{.95}
            \Block[l]{1-8}{\textbf{ECSpell} \textit{Odw}} &                    &            &            &                    &            &            &                    \\
            \midrule
            \texttt{CTG}                                  & \wm29.3            & \wm24.5    & \wm36.3    & \wm21.4            & \wm12.4    & \wm79.5    & \wm52.9            \\
            \midrule
            \texttt{OUR}                                  & \wm89.7            & \wm91.6    & \wm87.8    & \wm93.0            & \wm95.3    & \wm90.8    & \wm\wz1.3          \\
            \hdashedline
            \ \ \texttt{-\,DT}                            & \wz\ewm4.0         & \wz\ewm4.6 & \wz\ewm3.4 & \wz\ewm3.9         & \wz\ewm5.8 & \wz\ewm2.2 & \wz\wm0.0          \\
            \ \ \texttt{-\,DT}$^\dagger$                  & \ewm16.3           & \ewm16.9   & \ewm15.7   & \ewm12.7           & \ewm14.5   & \ewm10.9   & \wz\ewp2.5         \\
            \bottomrule
        \end{NiceTabular}
    }
    \caption{
        Ablation results of distortion model on \texttt{Baichuan2\;7B}.
        ``\texttt{CTG}'' means constrained text generation.
        ``\texttt{-DT}'' represents that we do not distinguish \texttt{Same Pinyin}, \texttt{Similar\,Pinyin}, and \texttt{Similar\,Shape}, and treat them as \texttt{Related} distortion.
        ``\texttt{-DT}$^\dagger$'' represents using the confusion set from \citet{wang-etal-2018-hybrid} to identify the \texttt{Related} distortion.
    }
    \label{tab:ablation:distortion}
\end{table}

To investigate the effectiveness of the minimal distortion model, we first remove the distortion model $p_{\texttt{DM}}(\boldsymbol{x}\mid \boldsymbol{y})$ from the decoding process.
Alternatively, we adopt a constrained text generation (CTG) approach to correct the input sentence.
For each step, we limit the vocabulary to tokens that are related to the corresponding characters in the input sentence,\!\footnote{Classified as \texttt{Identical}, \texttt{Same} \texttt{Pinyin}, \texttt{Similar} \texttt{Pinyin}, or \texttt{Similar} \texttt{Shape}.} and let the model select the most likely token from the constrained vocabulary.
The results are shown in the ``\texttt{CTG}'' column in Table~\ref{tab:ablation:distortion}.
We can see that the CTG performs poorly on all datasets.
This is because a Chinese character may have many similar characters.
Without the distortion model, the model is prone to replacing the original character with a higher-frequency similar character, leading to a large number of errors.

Next, we investigate the impact of the distortion type by treating three types of related but not identical distortions as a single distortion type.
As shown in the ``\texttt{-\,DT}'' column in Table~\ref{tab:ablation:distortion}, the performance drops significantly but not as severely as when removing the distortion model.
This performance drop is mainly due to a decrease in precision.

We also examine the effectiveness of our rule-based tool for identifying related distortions.
We replace our rule-based tool with the confusion set from \citet{wang-etal-2018-hybrid} to identify the related distortion.
The results in the ``\texttt{-\,DT}$^\dagger$'' column in Table~\ref{tab:ablation:distortion} show that the confusion set from \citet{wang-etal-2018-hybrid} is less effective than our rule-based tool, leading to more severe performance degradation.

\begin{table}[tb!]
    \centering
    \scalebox{0.85}{
        \setlength{\tabcolsep}{2.25pt}%
        \renewcommand{\arraystretch}{0.92}
        \begin{NiceTabular}{l>{\columncolor{figure_light_blue!40}}ccc;>{\columncolor{figure_light_blue!40}}ccc;>{\columncolor{figure_light_red!6}}c}
            \toprule
            \rowcolor[gray]{1.0}
            \Block[l]{1-1}{\textbf{System}}               & \textbf{S-F}\bgood & S-P\bgood  & S-R\bgood          & \textbf{C-F}\bgood & C-P\bgood  & C-R\bgood          & \textbf{FPR}\sgood \\
            \midrule
            \rowcolor[gray]{.95}
            \Block[l]{1-8}{\textbf{rSighan} \textit{15}}  &                    &            &                    &                    &            &                    &                    \\
            \midrule
            \texttt{Vanilla}                              & \wm18.0            & \wm15.9    & \wm20.6            & \wm20.7            & \wm14.3    & \wm37.6            & \wm52.9            \\
            \hdashedline
            \ \ \texttt{w/\,LR}                           & \ewp39.4           & \ewp43.4   & \ewp35.0           & \ewp43.7           & \ewp53.3   & \ewp23.9           & \ewm38.4           \\
            \ \ \texttt{w/\,FR}                           & \wz\ewp3.8         & \wz\ewp6.2 & \wz\ewp0.8         & \wz\ewp5.4         & \wz\ewp8.3 & \wz\uline{\ewm6.6} & \ewm19.3           \\
            \ \ \texttt{w/\,Both}                         & \ewp41.9           & \ewp50.1   & \ewp34.1           & \ewp47.4           & \ewp63.5   & \ewp23.0           & \ewm44.8           \\
            \midrule
            \midrule
            \rowcolor[gray]{.95}
            \Block[l]{1-8}{\textbf{Lemon} \textit{Nov}}   &                    &            &                    &                    &            &                    &                    \\
            \midrule
            \texttt{Vanilla}                              & \wm19.4            & \wm18.0    & \wm20.9            & \wm23.6            & \wm17.1    & \wm38.3            & \wm38.5            \\
            \hdashedline
            \ \ \texttt{w/\,LR}                           & \ewp17.1           & \ewp19.5   & \ewp14.6           & \ewp19.0           & \ewp21.9   & \wz\ewp8.6         & \ewm13.7           \\
            \ \ \texttt{w/\,FR}                           & \wz\ewp9.0         & \ewp13.5   & \wz\ewp4.7         & \wz\ewp8.5         & \ewp13.5   & \wz\uline{\ewm4.5} & \ewm18.8           \\
            \ \ \texttt{w/\,Both}                         & \ewp23.9           & \ewp34.2   & \ewp16.0           & \ewp24.1           & \ewp38.4   & \wz\ewp3.6         & \ewm25.0           \\
            \midrule
            \midrule
            \rowcolor[gray]{.95}
            \Block[l]{1-8}{\textbf{ECSpell} \textit{Odw}} &                    &            &                    &                    &            &                    &                    \\
            \midrule
            \texttt{Vanilla}                              & \wm65.3            & \wm65.3    & \wm65.3            & \wm70.4            & \wm65.4    & \wm76.2            & \wm10.1            \\
            \hdashedline
            \ \ \texttt{w/\,LR}                           & \ewp25.4           & \ewp26.9   & \ewp24.0           & \ewp22.5           & \ewp28.5   & \ewp15.6           & \wz\ewm9.7         \\
            \ \ \texttt{w/\,FR}                           & \wz\ewp4.7         & \ewp11.2   & \wz\uline{\ewm0.8} & \wz\ewp7.5         & \ewp19.7   & \wz\uline{\ewm4.5} & \wz\ewm6.7         \\
            \ \ \texttt{w/\,Both}                         & \ewp24.4           & \ewp26.4   & \ewp22.5           & \ewp22.6           & \ewp29.9   & \ewp14.6           & \wz\ewm8.8         \\
            \bottomrule
        \end{NiceTabular}
    }
    \caption{
        Ablation results of \texttt{Baichuan2\;7B}.
        ``\texttt{LR}'' and ``\texttt{FR}'' represent ``length reward'' and ``faithfulness reward'' respectively.
        ``\texttt{Both}'' means using both length reward and faithfulness reward.
    }
    \label{tab:ablation:results}
\end{table}

\subsection{Impact of Two Rewards}
In this work, we propose two rewards to optimize the decoding process: the length reward and the faithfulness reward.
The ablation study results of the two rewards are shown in Table~\ref{tab:ablation:results}.

The results show that the length reward significantly improves performance on all three datasets.
This improvement can be attributed to increases in both precision and recall, indicating that the length reward is crucial to our approach.
The faithfulness reward mainly contributes to improving precision, and it may slightly reduce recall.
Overall, the faithfulness reward balances the trade-off between precision and recall, leading to a higher $F_1$ score.

The combination of the two rewards can achieve better performance than using them separately, especially when datasets contain less formal text, more colloquial expressions, and more diverse named entities.

\subsection{Does Our Approach Work Well on Simpler LMs?}
\label{app:simpler_lm}
Though our primary focus is on the performance of our approach on LLMs, the language model term of Equation~\ref{eq:ncm_token} can be substituted with simpler models, such as $n$-gram models, masked language models, or small-scale causal language models.
In this subsection, we investigate the performance of our approach using these simpler language models.

The LMs we investigate include:
$n$\textbf{-gram LM}: \texttt{KLM},\footnote{\texttt{shibing624/chinese-kenlm-klm}} a 5-gram language model trained on the Chinese Gigaword corpus;
\textbf{Masked LM}: \texttt{BERT},\footnote{\texttt{bert-base-chinese}} a bidirectional language model pre-trained using the mask filling task and next sentence prediction task;
\textbf{Small causal LM}: \texttt{GPT2}\rlap{,}\footnote{\texttt{uer/gpt2-chinese-cluecorpussmall}} a small-scale causal language model (about 102M parameters) trained on the CLUECorpusSmall (about 5B characters).

The results are shown in Table~\ref{tab:ablation:simpler_lm}.
From these results, we can see that our approach also works with simpler LMs.
In the ECSpell-Odw dataset, our approach enables simpler language models (LMs) to achieve sentence- and character-level correction F1 scores higher than 50\% and 60\%, respectively.
However, the performance of our approach on simpler LMs still lags significantly behind that of the large language models (LLMs), highlighting the importance of the scale of pre-training data and model size.
\begin{table}[tb!]
    \setlength{\tabcolsep}{2.0pt}
    \centering
    \scalebox{0.83}{
        \begin{NiceTabular}{lcc>{\columncolor{figure_light_red!6}}c|cc>{\columncolor{figure_light_red!6}}c|cc>{\columncolor{figure_light_red!6}}c}
            \toprule
            \rowcolor[gray]{1.0}
            \Block[l]{2-1}{\textbf{System}} & \Block[c]{1-3}{\textbf{rSighan} \textit{15}} &           &           & \Block[c]{1-3}{\textbf{Lemon} \textit{Nov}} &           &           & \Block[c]{1-3}{\textbf{ECSpell} \textit{Odw}} &           &           \\
            \rowcolor[gray]{1.0}
                                            & S-F\bgood                                    & C-F\bgood & FPR\sgood & S-F\bgood                                   & C-F\bgood & FPR\sgood & S-F\bgood                                     & C-F\bgood & FPR\sgood \\
            \midrule
            \texttt{BC2\,13B}               & 59.6                                         & 67.3      & \wz8.3    & 43.5                                        & 47.9      & 13.0      & 92.0                                          & 93.8      & \wz0.4    \\
            \texttt{Q1\!.\!5\,14B}          & 57.6                                         & 66.0      & 10.2      & 36.4                                        & 42.6      & 21.2      & 87.4                                          & 91.6      & \wz2.9    \\
            \texttt{IL2\,20B}               & 60.5                                         & 67.8      & \wz8.3    & 40.5                                        & 45.3      & 15.1      & 91.1                                          & 93.8      & \wz0.4    \\
            \midrule
            \texttt{KLM}
                                            & 29.3                                         & 38.9      & 33.8      & \wz5.8                                      & 9.4       & 65.8      & 58.3                                          & 65.3      & 23.5      \\
            \texttt{BERT\,110M}
                                            & 31.3                                         & 34.0      & \wz0.2    & 13.3                                        & 12.5      & \wz0.6    & 59.1                                          & 63.6      & \wz0.0    \\
            \texttt{GPT2\,102M}
                                            & 55.0                                         & 64.7      & \wz8.1    & 26.1                                        & 30.8      & 28.4      & 78.6                                          & 85.0      & \wz5.4    \\
            \bottomrule
        \end{NiceTabular}
    }
    \caption{
        Results of applying our approach to simpler LMs.
    }
    \label{tab:ablation:simpler_lm}
\end{table}

\subsection{How to Introduce New Knowledge into Our Approach?}
\label{app:new_knowledge}

The LLM part of our approach offers a straightforward way to incorporate new knowledge \textbf{without the need for further training}, by \textbf{adding some text that describes the new knowledge as an input prefix}.

Given the new knowledge $\boldsymbol{k}$, Equation~\ref{eq:ncm_token} can be adjusted from $p(\boldsymbol{x}, \boldsymbol{y})$ to $p(\boldsymbol{x}, \boldsymbol{y}\!\mid\!\boldsymbol{k})$. We then have:
\begin{equation}
    \begin{aligned}
        p(\boldsymbol{x}, \boldsymbol{y} \mid \boldsymbol{k}) & = p(\boldsymbol{x} \mid \boldsymbol{y}, \boldsymbol{k}) \, p(\boldsymbol{y} \mid \boldsymbol{k})                     \\
                                                              & \approx p_{\texttt{DM}}(\boldsymbol{x} \mid \boldsymbol{y}) \, p_{\texttt{LLM}}(\boldsymbol{y} \mid \boldsymbol{k}),
    \end{aligned}
    \label{eq:ncm_token_new_knowledge}
\end{equation}
where, by assuming $\boldsymbol{x}$ and $\boldsymbol{k}$ are conditionally independent given $\boldsymbol{y}$, we approximate $p(\boldsymbol{x} \mid \boldsymbol{y}, \boldsymbol{k})$ as $p_{\texttt{DM}}(\boldsymbol{x} \mid \boldsymbol{y})$.
The second term, $p_{\texttt{LLM}}(\boldsymbol{y} \mid \boldsymbol{k})$, can be calculated by the LLM using the input prefix $\boldsymbol{k}$.

To illustrate this point, we conducted a simple experiment introducing domain and text format information as new knowledge into our approach.
We chose the \textbf{MCSCSet} for this experiment, as the sentences in it share a common characteristic: they are \textit{questions from patients}.
We can introduce this knowledge into the LLM by adding a simple input prefix $\boldsymbol{k} = \text{``患者提问：''}$ (``\textit{A patient asks:}'').

The results in Table~\ref{tab:discussion:intro_new_knowledge} demonstrate that introducing new knowledge into the LLM by merely modifying the input prefix can significantly improve the model's performance on the CSC task.

\begin{table}[tb!]
    \centering
    \setlength{\tabcolsep}{6.8pt}
    \scalebox{0.85}{
        \begin{NiceTabular}{lp{1.5cm}cc>{\columncolor{figure_light_red!6}}c}
            \toprule
            \rowcolor[gray]{1.0}
            \Block[l]{1-2}{\textbf{System}}        &                                     & S-F\bgood & C-F\bgood & FPR\sgood \\
            \midrule
            \Block[l]{2-1}{\texttt{Finetuned BERT}}
                                                   & \texttt{ORI}                        & 35.3      & 48.5      & \wm7.5    \\
                                                   & \quad \texttt{w/}\,$\boldsymbol{k}$ & \ewp0.7   & \ewp1.7   & \ewp0.1   \\
            \hdashedline
            \Block[l]{2-1}{\texttt{Softmasked BERT}}
                                                   & \texttt{ORI}                        & 35.3      & 48.5      & \wm8.1    \\
                                                   & \quad \texttt{w/}\,$\boldsymbol{k}$ & \ewp1.2   & \ewp2.1   & \ewm0.5   \\
            \hdashedline
            \Block[l]{2-1}{\texttt{ReLM}}
                                                   & \texttt{ORI}                        & 37.8      & 50.2      & \wm6.8    \\
                                                   & \quad \texttt{w/}\,$\boldsymbol{k}$ & \ewp0.9   & \ewp1.9   & \ewm0.2   \\
            \midrule
            \Block[l]{2-1}{\texttt{Baichuan2 13B}} & \texttt{OUR}                        & 66.0      & 76.9      & \wm1.7    \\
                                                   & \quad \texttt{w/}\,$\boldsymbol{k}$ & \ewp5.1   & \ewp5.4   & \ewm0.2   \\
            \hdashedline
            \Block[l]{2-1}{\texttt{Qwen1.5 14B}}   & \texttt{OUR}                        & 61.1      & 72.6      & \wm3.1    \\
                                                   & \quad \texttt{w/}\,$\boldsymbol{k}$ & \ewp9.1   & \ewp8.8   & \ewm1.0   \\
            \hdashedline
            \Block[l]{2-1}{\texttt{InternLM2 20B}} & \texttt{OUR}                        & 63.2      & 72.9      & \wm2.6    \\
                                                   & \quad \texttt{w/}\,$\boldsymbol{k}$ & \ewp4.8   & \ewp5.4   & \ewm0.0   \\
            \bottomrule
        \end{NiceTabular}
    }
    \caption{
        The results of introducing new knowledge by adding a prefix $\boldsymbol{k}$ to the input on the MCSCSet. ``\texttt{ORI}'' denotes the original input without any prefix.
    }
    \label{tab:discussion:intro_new_knowledge}
\end{table}

\vspace{0.5em}

We provide a real case from the MCSCSet to explain why this method works.

Consider the sentence ``未挨前兆'' (wèi āi qián zhào, ``\textit{without being near any prior warnings}''), which should be corrected to ``胃癌前兆'' (wèi ái qián zhào, ``\textit{early symptoms of stomach cancer}'') in the medical domain.
This sentence contains only four characters, insufficient to provide enough context for accurate correction, even for humans.

CSC models fail to correct this sentence or suggest incorrect corrections, such as ``未提前兆'' (wèi tí qián zhào, ``\textit{did not provide prior warnings}'') or ``未按前兆'' (wèi àn qián zhào, ``\textit{not according to the prior warnings}'').
However, if we add the prefix ``患者提问：'' (``\textit{A patient asks:}''), which provides the knowledge that the sentence is a patient's question about a medical condition, the model can correctly predict ``胃癌前兆''.

In addition to this simple experiment, we also provide an experiment in Appendix~\ref{app:context_as_new_knowledge} to show that we can use the context as new knowledge to improve the performance of the CSC model in real-world applications.

\subsection{More Discussions}
Due to space constraints, some interesting discussions have been moved to the Appendix.
These include:
a discussion on how the pre-training data of the LLM affects the performance (\ref{app:pre_training_data});
a comparison between our approach and the SFT approach (\ref{app:sft_comparison});
an analysis of the influence of beam size on the performance (\ref{app:beam_size});
an exploration of whether the imperfect estimation of the distortion model impacts the performance (\ref{app:distortion_analysis});
and a brief runtime analysis (\ref{app:inference_speed}).

    \section{Related Works}
\label{sec:related_work}

\subsection{Chinese Spelling Check}
Previous research on the CSC task can be divided into three eras, accompanied with paradigm shift.

\paragraph{The Early Unsupervised Era}
Early CSC approaches mainly utilized unsupervised pipeline systems \cite{yeh-etal-2013-chinese,yu-etal-2014-overview,yu-li-2014-chinese,huang-etal-2014-chinese,xie-etal-2015-chinese}.
These systems typicaly act in three main steps: error detection, candidate correction generation from a confusion set, and candidate ranking using a statistical $n$-gram language model.

\paragraph{The Supervised Learning Era}
By 2018, the advent of techniques for automatically generating pseudo-labeled data had begun to address the challenge of data scarcity in CSC \cite{wang-etal-2018-hybrid}, marking a shift in the paradigm of CSC research towards a supervised learning era dominated by deep neural networks.
This era saw researchers exploring various avenues to enhance CSC performance.
Some focused on finding better model architectures \cite{zhang-etal-2020-spelling,zhu-etal-2022-mdcspell}, while others delved into more effective training strategies \cite{liu-etal-2022-craspell,wu-etal-2023-rethinking,liu-etal-2023-chinese}.
Additionally, there was an effort to enrich models with information beyond text, such as phonetic or visual features \cite{cheng-etal-2020-spellgcn,xu-etal-2021-read,li-etal-2022-improving-chinese,liang-etal-2023-disentangled}.

Similar to our work, \citet{wu-etal-2023-rethinking} also decomposed $p(\boldsymbol{x}\mid\boldsymbol{y})$ into two parts to improve CSC performance.
However, they achieved this by adding an \emph{auxiliary training loss}.
Our work stands out by using an off-the-shelf LLM as the backbone and a minimal distortion model to achieve good CSC performance without any additional training.

\paragraph{The Era of LLMs}
Our work represents an initial foray into what can be considered the third era of CSC research: the era of LLMs.
This phase explores the potential of LLMs in addressing the CSC task.
As discussed in the introduction, related studies in this era fall into two main categories: \emph{prompt-based} and \emph{supervised fine-tuning}.
\citet{li-etal-2023-ineffectiveness} were the first to investigate the prompt-based approach under various settings.
Building on this work, \citet{dong-etal-2024-rich} proposed enriching prompts with additional information, such as pronunciation and character glyphs.
Compared to the prompt-based approach, the SFT-based approach has been shown to be more effective \cite{li-etal-2023-ineffectiveness}.
However, the performance of SFT-based LLMs still falls significantly behind pre-LLM methods.
\citet{li-etal-2024-cllm} argue that this underperformance is due to the mixed character-word tokenization used by LLMs for Chinese text.
To address this issue, they suggest replacing mixed tokenization with character-level tokenization before training LLMs on the CSC dataset.

In contrast to these methods, our approach requires neither prompts nor additional training.

\subsection{Decoding Methods of LLMs}
Intervening in the decoding process is a common approach to improve LLMs' task-specific performance.
There are two popular approaches in this category: \emph{Contrastive decoding} and \emph{Constrained decoding}.
Contrastive decoding \cite{li-etal-2023-contrastive} refines the output probabilities by comparing the output probabilities of expert and amateur models \cite{o-brien-lewis-2023-contrastive,shi-etal-2023-trusting}.
Constrained decoding, on the other hand, uses constraints to guide the decoding process, making the output more aligned with the task-specific requirements \cite{wang-etal-2023-grammar,geng-etal-2023-grammar}.

Our work is closely related to the constrained decoding approaches, where a distortion model is used to influence the LLM decoding process.

    \section{Conclusion}
In this work, we propose a simple, training-free, and prompt-free approach to leverage LLMs for the CSC task.
Two components, a large language model and a minimal distortion model, co-operate to correct spelling errors.
We alleviate the local optima problem and over-correction issue, with two simple strategies, length reward and faithfulness reward, respectively.
Our comprehensive experiments have shown that our approach significantly improves LLM performance.
Through our approach, LLMs demonstrate remarkable domain generalization capabilities, surpassing SOTA domain-general CSC models, that are trained on extensive synthetic CSC data, on most datasets.

    \section*{Limitations}
\paragraph{Feasibility}
The scope of this study is limited to the task of Chinese spelling correction, which is a subset of text error correction.
Most of our design choices are tailored to the characteristics of Chinese and the specific requirements of the CSC task.

However, our approach has the potential to be directly applied to some other languages.
\begin{CJK*}{UTF8}{min}%
    For example, in Japanese and Korean, we can also categorize errors into phonetic similarities,
    such as (や, \textit{ya})-(な, \textit{na}) in Japanese
\end{CJK*}%
\begin{CJK*}{UTF8}{mj}%
    or (후, \textit{hu})-(부, \textit{bu}) in Korean, and shape similarities,
\end{CJK*}%
\begin{CJK*}{UTF8}{min}%
    like (ュ, \textit{yu})-(ェ, \textit{e}) in Japanese.
\end{CJK*}%
For languages using a phonetic writing system, like English, minor adjustments such as adding \texttt{INSERT}, \texttt{DELETE}, and \texttt{REORDER} operations will be sufficient to make it work.

Comparatively, handling complex text errors that involve grammar, semantics, or pragmatics, are more challenging.
To tackle these errors, one could design an appropriate distortion model, though it might necessitate the adoption of more intricate rules or the implementation of a model based on neural networks.
In our future work, we aim to explore ways that would allow our approach to handle these complex errors.

\paragraph{Computational Cost}
Our approach requires the use of LLMs, which introduces additional computational costs.
However, many existing techniques, such as quantization \cite{frantar-etal-2022-gptq,lin-etal-2024-awq}, pruning \cite{ma-etal-2023-llm-pruner,zhu-etal-2024-llama-moe}, distillation \cite{hsieh-etal-2023-distilling}, and more efficient framework implementations \cite{dao-etal-2023-flashattention-2,yang-etal-2024-parallelizing}, can be directly applied to our method to reduce these costs.

    \ifarxiv%
        \section*{Acknowledgements}
First and foremost, we would like to express our deepest gratitude to all anonymous reviewers for their invaluable time and constructive comments on our paper.
We would also like to thank Chen Gong, Tong Zhu, Shilin Zhou, and Yu Zhang for their help in polishing our paper.

This work was supported by National Natural Science Foundation of China (Grant No. 62176173 and 62261160648), Alibaba Group through Alibaba Innovative Research Program, and a Project Funded by the Priority Academic Program Development (PAPD) of Jiangsu Higher Education Institutions.

    \fi%

    \bibliography{custom}

    \clearpage
    \appendix
\section{Special Acknowledgements}

We would like to extend our special thanks to all anonymous reviewers for their valuable comments and suggestions.

\begin{asparaitem}[\textbf{Reviewer}]
    \item \textit{gUCq} highlighted unclear descriptions and missing experiments, such as the comparison with simpler LMs, in our initial version of the paper. The revisions made in response to these suggestions have significantly improved the quality of our work.
    \item \textit{B44m} provided strong positive feedback on our work while also identifying missing details in the experimental setup, the absence of results on few-shot settings with GPT-4, and other aspects. Addressing these points made our paper more comprehensive and rigorous.
    \item \textit{cWnK} raised concerns about the flexibility of introducing new knowledge. This insightful comment motivated us to further explore the topic and provide a simple solution in \S\ref{app:new_knowledge}.
    \item \textit{cnvj}, \textit{wY4T}, and \textit{QXDJ} gave our work high evaluations and provided numerous constructive comments and suggestions. Their recognition encourages us to continue refining our paper.
\end{asparaitem}

\section{Implement of Distortion Model}
\label{app:detailed_type}

\subsection{Standard of Transformation Types}

\paragraph{Identical Transformations}
An identical distortion occurs when the input character is the same as the correct character.

\paragraph{Same Pinyin}
Characters that share the same pronunciation, disregarding tone, undergo a ``\texttt{Same Pinyin}'' distortion.
Due to the existence of heteronyms in Chinese, such as ``和'', which can be pronounced in multiple ways including ``\textit{hé}'', ``\textit{hè}'', ``\textit{huó}'', ``\textit{huò}'', and ``\textit{hú}'', we classify two characters as undergoing a same pinyin distortion if they share at least one pronunciation.
The \texttt{pypinyin}\footnote{\url{https://github.com/mozillazg/python-pinyin}} library is utilized to determine character pronunciations, with the \texttt{ktghz2013} and \texttt{large\_pinyin} from \texttt{pypinyin-dict}\footnote{\url{https://github.com/mozillazg/pypinyin-dict}} providing a more accurate pronunciation for these determinations.

\paragraph{Similar Pinyin}
We categorize distortions as ``\texttt{Similar Pinyin}'' when two characters have pronunciation that is recognized as similar by predefined rules, which are based on \citet{yang-etal-2023-chinese}.
For instance, `\textit{qī}'' and ``\textit{jī}'' are considered similar due to the common mispronunciation of the consonant ``\textit{q}'' as ``\textit{j}''.
A  list of consonants and vowels considered similar can be found in Tables~\ref{tab:consonant_similar} and~\ref{tab:vowel_similar}, respectively.

\paragraph{Similar Shape}
The similarity in the shape of characters is evaluated by combining their four-corner code with their radical and component information.
For example, the characters ``机'' and ``仉'' have the four-corner codes ``\texttt{4\textcolor{figure_blue}{\textbf{7}}9\textcolor{figure_blue}{\textbf{1}}0}'' and ``\texttt{2\textcolor{figure_blue}{\textbf{7}}2\textcolor{figure_blue}{\textbf{1}}0}'', respectively.
Given that the last digit primarily serves to distinguish characters with identical preceding digits and that ``机'' and ``仉'' share two of these digits, their four-corner code similarity is calculated as $2 \times \frac{1}{4} = 0.5$.
Considering their radical and component (``木, 几'' for ``机'' and ``人, 几'' for ``仉''), which share the component ``几'' but differ in radicals, their similarity is $1 \times \frac{1}{2} = 0.5$.
Thus, the overall similarity is averaged to 0.5.
With a similarity threshold set at 0.45, these characters are considered to undergo a similar shape distortion.
Furthermore, character pairs where one is a radical or component of the other, such as ``机'' and ``几'', are also classified under similar shape distortions.

\vspace{1em}

All non-Chinese characters are only allowed to be transformed into themselves.

\begin{table}[tb!]
    \newcommand\forbid{\rlap{\textcolor{black!10}{\ding{110}}}\textcolor{black!18}{\ding{53}}}
    \newcommand\forbidr{\rlap{\textcolor{black!30}{\ding{110}}}\textcolor{black!38}{\ding{53}}}
    \newcommand\self{\textcolor{figure_green!10}{\ding{108}}}
    \setlength{\tabcolsep}{4.0pt}
    \centering
    \begin{NiceTabular}{lcccccccccc}
        \toprule
        \Block[c]{1-11}{\textbf{Corrected} $\rightarrow$ \textbf{Input}} &               &            &            &            &            &            &            &             &             &             \\
        \midrule
        \textit{j}                                                       & $\rightarrow$ & \self      & \textit{q} & \textit{x} & \textit{z} & \forbid    & \forbid    & \forbid     & \forbid     & \forbid     \\
        \textit{q}                                                       & $\rightarrow$ & \textit{j} & \self      & \textit{x} & \forbid    & \textit{c} & \forbid    & \forbid     & \forbid     & \forbid     \\
        \textit{x}                                                       & $\rightarrow$ & \textit{j} & \textit{q} & \self      & \forbid    & \forbid    & \textit{s} & \forbid     & \forbid     & \forbid     \\
        \textit{z}                                                       & $\rightarrow$ & \textit{j} & \forbid    & \forbid    & \self      & \textit{c} & \textit{s} & \textit{zh} & \forbid     & \forbid     \\
        \textit{c}                                                       & $\rightarrow$ & \forbid    & \textit{q} & \forbid    & \textit{z} & \self      & \textit{s} & \forbid     & \textit{ch} & \forbid     \\
        \textit{s}                                                       & $\rightarrow$ & \forbid    & \forbid    & \forbidr   & \textit{z} & \textit{c} & \self      & \forbid     & \forbid     & \textit{sh} \\
        \textit{zh}                                                      & $\rightarrow$ & \forbid    & \forbid    & \forbid    & \textit{z} & \forbid    & \forbid    & \self       & \textit{ch} & \textit{sh} \\
        \textit{ch}                                                      & $\rightarrow$ & \forbid    & \forbid    & \forbid    & \forbid    & \textit{c} & \forbid    & \textit{zh} & \self       & \textit{sh} \\
        \textit{sh}                                                      & $\rightarrow$ & \forbid    & \forbid    & \forbid    & \forbid    & \forbid    & \textit{s} & \textit{zh} & \textit{ch} & \self       \\
        \midrule
        \textit{r}                                                       & $\rightarrow$ & \self      & \textit{l} & \forbid    & \forbid    & \forbid    & \forbid    & \forbid     & \forbid                   \\
        \textit{l}                                                       & $\rightarrow$ & \textit{r} & \self      & \textit{n} & \textit{d} & \textit{t} & \forbid    & \forbid     & \forbid                   \\
        \textit{n}                                                       & $\rightarrow$ & \forbid    & \textit{l} & \self      & \textit{d} & \textit{t} & \forbid    & \forbid     & \forbid                   \\
        \textit{d}                                                       & $\rightarrow$ & \forbid    & \textit{l} & \textit{n} & \self      & \textit{t} & \textit{b} & \forbid     & \forbid                   \\
        \textit{t}                                                       & $\rightarrow$ & \forbid    & \textit{l} & \textit{n} & \textit{d} & \self      & \forbid    & \textit{p}  & \forbid                   \\
        \textit{b}                                                       & $\rightarrow$ & \forbid    & \forbid    & \forbid    & \textit{d} & \forbid    & \self      & \textit{p}  & \textit{m}                \\
        \textit{p}                                                       & $\rightarrow$ & \forbid    & \forbid    & \forbid    & \forbid    & \textit{t} & \textit{b} & \self       & \forbidr                  \\
        \textit{m}                                                       & $\rightarrow$ & \forbid    & \forbid    & \forbid    & \forbid    & \forbid    & \textit{b} & \textit{p}  & \self                     \\
        \midrule
        \textit{g}                                                       & $\rightarrow$ & \self      & \textit{k} & \textit{h} & \forbid                                                                        \\
        \textit{k}                                                       & $\rightarrow$ & \textit{g} & \self      & \textit{h} & \forbid                                                                        \\
        \textit{h}                                                       & $\rightarrow$ & \textit{g} & \textit{k} & \self      & \textit{f}                                                                     \\
        \textit{f}                                                       & $\rightarrow$ & \forbid    & \forbid    & \textit{h} & \self                                                                          \\
        \bottomrule
    \end{NiceTabular}
    \caption{
        Consonants with similar pronunciation.
    }
    \label{tab:consonant_similar}
\end{table}

\begin{table}[tb!]
    \newcommand\forbid{\rlap{\textcolor{black!10}{\ding{110}}}\textcolor{black!18}{\ding{53}}}
    \newcommand\self{\textcolor{figure_green!10}{\ding{108}}}
    \setlength{\tabcolsep}{4.0pt}
    \centering
    \begin{NiceTabular}{lccccccc}
        \toprule
        \Block[c]{1-8}{\textbf{Corrected} $\rightarrow$ \textbf{Input}} &               &             &              &              &               &              &               \\
        \midrule
        \textit{an}                                                     & $\rightarrow$ & \self       & \textit{ang} & \textit{uan} & \textit{uang} & \textit{ian} & \forbid       \\
        \textit{ang}                                                    & $\rightarrow$ & \textit{an} & \self        & \textit{uan} & \textit{uang} & \forbid      & \textit{iang} \\
        \textit{uan}                                                    & $\rightarrow$ & \textit{an} & \textit{ang} & \self        & \textit{uang} & \textit{ian} & \forbid       \\
        \textit{uang}                                                   & $\rightarrow$ & \textit{an} & \textit{ang} & \textit{uan} & \self         & \forbid      & \textit{iang} \\
        \textit{ian}                                                    & $\rightarrow$ & \textit{an} & \forbid      & \textit{uan} & \forbid       & \self        & \textit{iang} \\
        \textit{iang}                                                   & $\rightarrow$ & \forbid     & \textit{ang} & \forbid      & \textit{uang} & \textit{ian} & \self         \\
        \midrule
        \textit{en}                                                     & $\rightarrow$ & \self       & \textit{eng} & \textit{un}  & \forbid                                      \\
        \textit{eng}                                                    & $\rightarrow$ & \textit{en} & \self        & \forbid      & \forbid                                      \\
        \textit{un}                                                     & $\rightarrow$ & \textit{en} & \forbid      & \self        & \textit{ong}                                 \\
        \textit{ong}                                                    & $\rightarrow$ & \forbid     & \forbid      & \textit{un}  & \self                                        \\
        \midrule
        \textit{in}                                                     & $\rightarrow$ & \self       & \textit{ing}                                                               \\
        \textit{ing}                                                    & $\rightarrow$ & \textit{in} & \self                                                                      \\
        \midrule
        \textit{o}                                                      & $\rightarrow$ & \self       & \textit{uo}                                                                \\
        \textit{uo}                                                     & $\rightarrow$ & \textit{o}  & \self                                                                      \\
        \midrule
        \textit{ü}                                                      & $\rightarrow$ & \self       & \textit{u}                                                                 \\
        \textit{u}                                                      & $\rightarrow$ & \textit{ü}  & \self                                                                      \\
        \bottomrule
    \end{NiceTabular}
    \caption{
        Vowels with similar pronunciation.
    }
    \label{tab:vowel_similar}
\end{table}

\begin{table*}[tb!]
    \centering
    \setlength{\tabcolsep}{4.0pt}
    \begin{NiceTabular}{p{14.8em}ccc|c|c|ccc}
        \toprule
        \rowcolor[gray]{.9}
        \textbf{Datasets}                                        & \Block[c]{1-3}{\textbf{rSighans}} &              &              & \Block[c]{1-1}{\textbf{CSCD}} & \Block[c]{1-1}{\textbf{MCSC}} & \Block[c]{1-3}{\textbf{ECSpell}} &              &              \\
        \textbf{Subsets}                                         & \textit{Y13}                      & \textit{Y14} & \textit{Y15} & \textit{Test}                 & \textit{Test}                 & \textit{Law}                     & \textit{Med} & \textit{Odw} \\
        \midrule
        \textbf{\#Sentence}                                      & 1,000                             & 1,062        & 1,100        & 5,000                         & 19,650                        & 500                              & 500          & 500          \\
        \textbf{Erroneous Sentence Ratio}                        & 97.70                             & 56.69        & 56.18        & 46.06                         & 50.00                         & 51.00                            & 45.20        & 52.40        \\
        \textbf{Average Length}                                  & 74.33                             & 50.01        & 30.64        & 57.63                         & 10.91                         & 29.74                            & 49.60        & 40.51        \\
        \textbf{Average Error/Sentence}                          & \wz1.48                           & \wz0.88      & \wz0.78      & \wz0.51                       & \wz0.93                       & \wz0.78                          & \wz0.71      & \wz0.81      \\
        \midrule
        \rowcolor[gray]{.95}
        \Block[l]{1-5}{\textbf{Distortion Type Proportion (\%)}} &                                   &              &              &                               &                               &                                  &              &              \\
        \quad \texttt{Identical}                                 & 98.01                             & 98.25        & 97.45        & 99.12                         & 91.47                         & 97.38                            & 98.56        & 98.01        \\
        \hdashedline
        \quad \texttt{Same Pinyin}                               & \wz1.62                           & \wz1.30      & \wz1.83      & \wz0.74                       & \wz6.60                       & \wz1.82                          & \wz1.15      & \wz1.55      \\
        \quad \texttt{Similar Pinyin}                            & \wz0.28                           & \wz0.40      & \wz0.66      & \wz0.13                       & \wz1.05                       & \wz0.51                          & \wz0.19      & \wz0.28      \\
        \quad \texttt{Similar Shape}                             & \wz0.05                           & \wz0.01      & \wz0.03      & \wz0.00                       & \wz0.39                       & \wz0.25                          & \wz0.08      & \wz0.13      \\
        \hdashedline
        \rowcolor[gray]{.97}
        \quad \texttt{Unrelated}                                 & \wz0.04                           & \wz0.04      & \wz0.02      & \wz0.00                       & \wz0.45                       & \wz0.04                          & \wz0.01      & \wz0.02      \\
        \midrule
        \textbf{Recall Upper Bound}                              & 97.24                             & 97.18        & 98.71        & 99.70                         & 90.82                         & 97.65                            & 98.67        & 98.47        \\
        \bottomrule
    \end{NiceTabular}
    \\[2pt]
    \setlength{\tabcolsep}{4.0pt}
    \begin{NiceTabular}{p{12.5em}ccccccc|c}
        \toprule
        \rowcolor[gray]{.9}
        \textbf{Datasets}                                        & \Block[c]{1-7}{\textbf{Lemon}} &              &              &              &              &              &              & \Block[c]{1-1}{\textbf{Pseudo-Dev}} \\
        \textbf{Subsets}                                         & \textit{Car}                   & \textit{Cot} & \textit{Enc} & \textit{Gam} & \textit{Med} & \textit{New} & \textit{Nov} & --                                  \\
        \midrule
        \textbf{\#Sentence}                                      & 3,410                          & 1,026        & 3,434        & 400          & 2,090        & 5,892        & 6,000        & 2,000                               \\
        \textbf{Erroneous Sentence Ratio}                        & 51.09                          & 46.20        & 50.99        & 38.75        & 50.38        & 50.00        & 50.23        & 93.55                               \\
        \textbf{Average Length}                                  & 43.44                          & 40.12        & 39.83        & 32.99        & 39.28        & 25.16        & 36.24        & 36.94                               \\
        \textbf{Average Error/Sentence}                          & \wz0.56                        & \wz0.47      & \wz0.52      & \wz0.41      & \wz0.49      & \wz0.55      & \wz0.57      & \wz1.42                             \\
        \midrule
        \rowcolor[gray]{.95}
        \Block[l]{1-5}{\textbf{Distortion Type Proportion (\%)}} &                                &              &              &              &              &              &              &                                     \\
        \quad \texttt{Identical}                                 & 98.64                          & 98.78        & 98.63        & 98.73        & 98.64        & 97.80        & 98.43        & 96.15                               \\
        \hdashedline
        \quad \texttt{Same Pinyin}                               & \wz0.90                        & \wz0.75      & \wz0.93      & \wz0.89      & \wz0.94      & \wz1.50      & \wz0.95      & \wz2.34                             \\
        \quad \texttt{Similar Pinyin}                            & \wz0.31                        & \wz0.25      & \wz0.28      & \wz0.26      & \wz0.27      & \wz0.51      & \wz0.43      & \wz0.78                             \\
        \quad \texttt{Similar Shape}                             & \wz0.02                        & \wz0.07      & \wz0.06      & \wz0.01      & \wz0.02      & \wz0.05      & \wz0.02      & \wz0.40                             \\
        \hdashedline
        \rowcolor[gray]{.97}
        \quad \texttt{Unrelated}                                 & \wz0.12                        & \wz0.14      & \wz0.09      & \wz0.11      & \wz0.12      & \wz0.13      & \wz0.16      & \wz0.31                             \\
        \midrule
        \textbf{Recall Upper Bound}                              & 91.38                          & 89.34        & 94.28        & 90.54        & 92.82        & 93.98        & 89.08        & 88.03                               \\
        \bottomrule
    \end{NiceTabular}
    \caption{
        The statistics of the datasets used in the experiments. \textbf{Recall Upper Bound} represents the sentence-level upper bound of the recall under the distortion model that we use in this work.
    }
    \label{tab:data_statistics}
\end{table*}

\subsection{Type Priority}
In scenarios where a character can be classified under multiple distortion types, for example, ``机'' (\textit{jī}) and ``玑'' (\textit{jī}), which can be classified as both having the same pinyin and a similar shape, we prioritize the distortion type according to the following order:
\begin{inparaenum}[1)]
    \item \texttt{Identical};
    \item \texttt{Same Pinyin};
    \item \texttt{Similar Pinyin};
    \item \texttt{Similar Shape};
    \item \texttt{Unrelated}.
\end{inparaenum}

\subsection{Using an Inverted Index for Efficient Distortion Model Calculation}

During each decoding step, the distortion model calculates the probability of transforming the input sequence $x_{a:b}$ into a candidate token $t_i$:
\begin{equation}
    g(x, t_i) = \sum_{r=1}^k \log p_{\mathtt{DM}}(c_r \mid  x_{l+r}),
\end{equation}
where the function $g(x, t_i)$ must be computed for each candidate token $t_i$ in the vocabulary $\mathcal{V}$, resulting in a huge computational cost.

To address this challenge, we propose the use of an inverted index to reduce the calculation process, by only considering relevant tokens, and ignoring irrelevant tokens.
For a token, we can pre-construct indexed entries to represent it, such as \texttt{<0,ji,SamePinyin>}, \texttt{<1,kou,SimilarPinyin>}, and \texttt{<0,仉,SimilarShape>} for ``机构'' (\textit{jī gòu}).
Upon receiving an input sequence, the index enables rapid retrieval of relevant tokens, thereby limiting probability calculations exclusively to these tokens.
As the subset of relevant tokens is substantially smaller than the complete token set, employing an inverted index considerably reduces the computational burden.

\subsection{Small Tricks for Distortion Model}
We adopt three small tricks to enhance our distortion model.
First, for character pairs commonly misused in everyday writing, such as ``的'', ``地'', and ``得'', we categorize these as ``\texttt{Identical}'' distortions, allowing the model to correct these errors with lower difficulty.

Second, we found that, although the previously described rules adequately cover most similar relationships between characters, a few exceptions, approximately 0.01\% of total character pairs, still persist.
To identify these outliers, we leveraged tools from previous studies \cite{wu-etal-2023-rethinking,hu-etal-2024-cscd} by incorporating their structure confusion sets and spelling similarity matrices.
We classify character pairs found within the structure confusion set or those with a spelling similarity matrix distance of less than 1 as ``\texttt{Other Similar}'' distortions.

Finally, we have chosen not to entirely exclude unrelated distortions.
Instead, we allow each token to possess up to one unrelated character distortion, to which we assign a very low probability ($\log p_{\mathtt{DM}} = -15$).

Employing these tricks has led to marginal yet consistent improvements in our approach's performance.

\section{Details of Experiments}
\label{app:experiments}
\subsection{Details of Real-world Test Sets}
This section details the test sets used in our study, providing insights into their composition and relevance to real-world Chinese text.
\label{app:test_sets}
\begin{asparaitem}[$\bullet$]
    \item \textbf{Sighan series}: This series of datasets is one of the most widely used benchmark datasets for Chinese spelling correction \cite{wu-etal-2013-chinese,yu-etal-2014-overview,tseng-etal-2015-introduction}. However, it faces criticism for two main reasons: firstly, it consists of essays written by Chinese learners, which may not accurately represent typical Chinese texts. Secondly, its limited diversity could hinder the evaluation of models' generalization capabilities. Despite these concerns, we include it in our evaluation to allow for comparison with prior studies. However, we utilize the revised version by \citet{yang-etal-2023-chinese}, which has manually verified and corrected the errors in the original dataset.
    \item \textbf{CSCD-NS}: A real-world Chinese social media corpus collected and annotated by \citet{hu-etal-2024-cscd}. It can better represent the variety of texts found in real-world settings and includes a broad spectrum of errors.
    \item \textbf{MCSCSet}: A large-scale corpus from the medical domain, collected and annotated by \citet{jiang-etal-2022-mcscset}. It features numerous errors specific to medical terminology, making it an excellent resource for evaluating models' generalization capabilities in this area.
    \item \textbf{ECSpell}: A small-scale, multi-domain corpus annotated by \citet{lv-etal-2023-ecspell}. It encompasses three domains: legal documents, medical treatments, and official document writing.
    \item \textbf{Lemon}: The most recent and largest multi-domain corpus to date, collected and annotated by \citet{wu-etal-2023-rethinking}. It spans seven domains: law, medicine, encyclopedia, gaming, automotive, contracts, news, and novels. The original dataset also includes sighan 15 as a subset, which we have considered as a part of the Sighan series and excluded from Lemon.
\end{asparaitem}

The detailed statistics of these datasets are shown in Table~\ref{tab:data_statistics}.

The recall upper bound in the statistics is obtained by calculating the number of sentences that can potentially be fully corrected out of the total number of sentences in the dataset.
A sentence has the potential to be fully corrected if all the distortion types between each pair of source and target characters can be categorized into \texttt{Identical}, \texttt{Same\,Pinyin}, \texttt{Similar\,Pinyin}, and \texttt{Similar\,Shape}.

\begin{figure*}[tb!]
    \captionsetup[subfigure]{skip=2pt}
    \centering%
    \newtcolorbox{promptbox}[1]{%
        left=0pt,
        right=0pt,
        top=0pt,
        bottom=0pt,
        boxsep=6pt,
        colback=black!3,
        colframe=black,
        title={#1},
    }

    \centering%
    \begin{minipage}[b]{0.98\linewidth}
        \begin{promptbox}{System and User Prompts for baselines}
            \textbf{\texttt{System Prompt:}}\\
            你是一个优秀的中文拼写纠错模型，中文拼写纠错模型即更正用户输入句子中的拼写错误。\\
            \textbf{\texttt{User Prompt:}}\\
            你需要识别并纠正用户输入的句子中可能的错别字并输出正确的句子，纠正时必须保证改动前后句子必须等长，在纠正错别字的同时尽可能减少对原句子的改动(不添加额外标点符号，不添加额外的字，不删除多余的字)。只输出没有错别字的句子，不要添加任何其他解释或说明。如果句子没有错别字，就直接输出和输入相同的句子。
        \end{promptbox}
    \end{minipage}
    \caption{
        Prompt templates used in our \texttt{FSP} and \texttt{ZSP} baselines.
    }
    \label{fig:prompt_examples}
\end{figure*}

\subsection{Implementation Details of Prompt-based Method}
\label{app:prompt_examples}
In this work, we use the prompt-based method to activate the CSC ability of the baseline LLMs.
The task-specific instructions are adopted from \citet{li-etal-2023-ineffectiveness}.
The prompt used for the baselines are shown in Figure~\ref{fig:prompt_examples}.
We disable the sampling mechanism and set the temperature to 0.0 to ensure deterministic decoding.
For few-shot prompting methods, where the example selection strategy involves random selection, we conduct three runs and report the average results.
The only exception is the \texttt{GPT4} model, which we run only once due to the high cost of using the model.

\subsection{Few-shot Examples Selection Strategy for Baselines}
\label{app:few_shot_examples}
\citet{li-etal-2023-ineffectiveness} proposed three selection strategies for CSC few-shot prompting methods:
\begin{inparaenum}[1)]
    \item \texttt{Random}: randomly select $m$ examples;
    \item \texttt{Balanced}: randomly select $m$ examples with a balanced distribution of correct and error examples;
    \item \texttt{Similarity}: select the $m$ most similar in-context examples for each input sentence using the \texttt{BM25} and \texttt{Rouge} similarity metrics.
\end{inparaenum}

They found that the performance of few-shot prompting depends on the selection of in-context examples.
Different selection strategies may lead to distinct results.
Among the three strategies, \texttt{Similarity} was found to be the most effective.

However, the \texttt{Similarity} strategy is not always the optimal choice.
In preliminary experiments, we observed that this strategy sometimes causes GPT family models to perform worse than the zero-shot prompting method.
Upon analyzing the results, we found that GPT models are particularly sensitive to discrepancies in the proportion of erroneous sentences between the few-shot prompting examples and the target data.
The examples selected using the \texttt{Similarity} strategy tend to have a similar proportion of erroneous sentences as the dataset used for selection.
In our work, we use Pseudo-Dev dataset to select few-shot prompting examples, which contains a higher proportion of erroneous sentences (87\%–94\%) compared to the target data (50\%–56\%).
This discrepancy causes the GPT models to be more aggressive in correcting errors.

To ensure the effectiveness of the few-shot prompting method, we conducted experiments to determine the optimal strategy for each LLM we used.
For open-source LLMs, which include both `\texttt{Base}' and `\texttt{Chat}' versions, we experimented with both versions and selected the best one for each LLM.
The final choice of selection strategy is shown in Table~\ref{tab:hyperparameters:few-shot}.

\begin{table}[tb!]
    \centering
    \begin{NiceTabular}{lcc}
        \toprule
        \rowcolor[gray]{1.0}
        \textbf{Model}          & \textbf{Version} & \textbf{Strategy}  \\
        \midrule
        \texttt{Baichuan2\;13B} & \texttt{Base}    & \texttt{Similariy} \\
        \texttt{Qwen1.5\;14B}   & \texttt{Base}    & \texttt{Balanced}  \\
        \texttt{InternLM2\;20B} & \texttt{Chat}    & \texttt{Similariy} \\
        \hdashedline
        \texttt{GPT3.5}         & --               & \texttt{Balanced}  \\
        \texttt{GPT4}           & --               & \texttt{Balanced}  \\
        \bottomrule
    \end{NiceTabular}
    \caption{
        The model version and examples selection strategy we used for few-shot baseline.
    }
    \label{tab:hyperparameters:few-shot}
\end{table}

\begin{table*}[tb!]
    \centering
    \setlength{\tabcolsep}{4.25pt}
    \begin{NiceTabular}{lcccc|ccc|ccccccc}
        \toprule
        \rowcolor[gray]{.9}
        \textbf{Datasets}                                                                                                   &              & \Block[c]{1-3}{\textbf{rSighans}} &               &               & \Block[c]{1-3}{\textbf{ECSpell}} &               &               & \Block[c]{1-7}{\textbf{Lemon}} &               &               &               &               &               &               \\
        \textbf{Subsets}                                                                                                    &              & \textit{Y13}                      & \textit{Y14}  & \textit{Y15}  & \textit{Law}                     & \textit{Med}  & \textit{Odw}  & \textit{Car}                   & \textit{Cot}  & \textit{Enc}  & \textit{Gam}  & \textit{Med}  & \textit{New}  & \textit{Nov}  \\
        \midrule
        \rowcolor[gray]{.95}
        \Block[l]{1-15}{\texttt{Domain-Specific\;SOTAs} (\textit{Trained on in-domain gold-standard data of each dataset})} &              &                                   &               &               &                                  &               &               &                                &               &               &               &               &               &               \\
        \Block[l]{1-2}{\texttt{ReaLiSe}}                                                                                    &              & 70.1                              & 64.0          & 73.9          & \ood{38.9}                       & \ood{23.1}    & \ood{42.8}    & \ood{32.5}                     & \ood{40.1}    & \ood{29.1}    & \ood{12.6}    & \ood{31.8}    & \ood{31.2}    & \ood{20.2}    \\
        \Block[l]{1-2}{\citet{liu-etal-2023-chinese}}                                                                       &              & --                                & --            & --            & 91.2                             & 82.4          & 83.6          & --                             & --            & --            & --            & --            & --            & --            \\
        \midrule
        \rowcolor[gray]{.95}
        \Block[l]{1-15}{\texttt{Domain-General\;SOTAs} (\textit{Trained on about 34M synthetic CSC data})}                  &              &                                   &               &               &                                  &               &               &                                &               &               &               &               &               &               \\
        \Block[l]{1-2}{\texttt{Finetuned\;BERT}}                                                                            &              & 50.6                              & 40.4          & 51.6          & 58.5                             & 47.8          & 65.1          & 52.0                           & 63.1          & 45.3          & 32.8          & 50.7          & 56.1          & 35.8          \\
        \Block[l]{1-2}{\texttt{Softmasked\;BERT}}                                                                           &              & \textbf{51.6}                     & 40.2          & 51.3          & 58.5                             & 48.5          & 65.9          & 52.3                           & 63.8          & 44.1          & 28.3          & 48.9          & 55.6          & \textbf{37.7} \\
        \Block[l]{1-2}{\texttt{ReLM}}                                                                                       &              & 45.8                              & \textbf{40.6} & \textbf{55.5} & \textbf{60.4}                    & \textbf{50.9} & \textbf{66.5} & \textbf{53.3}                  & \textbf{66.7} & \textbf{47.7} & \textbf{33.7} & \textbf{53.8} & \textbf{58.8} & 37.1          \\
        \midrule
        \rowcolor[gray]{.95}
        \Block[l]{1-15}{\texttt{LLMs} (\textit{without CSC-specific training})}                                             &              &                                   &               &               &                                  &               &               &                                &               &               &               &               &               &               \\
        \Block[c]{3-1}{\texttt{Baichuan2}                                                                                                                                                                                                                                                                                                                                                                          \\[-4pt]\texttt{\footnotesize{(13B)}}}
                                                                                                                            & \texttt{ZSP} & 26.4                              & 12.0          & 18.5          & 37.6                             & 23.0          & 43.0          & 15.3                           & 14.9          & 24.0          & 12.7          & 21.6          & 19.8          & 14.1          \\
                                                                                                                            & \texttt{FSP} & 41.1                              & 23.1          & 31.3          & 60.2                             & 50.4          & 60.0          & 32.2                           & 45.3          & 38.9          & 24.6          & 39.0          & 39.7          & 26.4          \\
                                                                                                                            & \texttt{OUR} & \textbf{63.6}                     & \textbf{54.1} & \textbf{59.6} & \textbf{82.6}                    & \textbf{78.9} & \textbf{92.0} & \textbf{52.7}                  & \textbf{62.9} & \textbf{51.9} & \textbf{37.1} & \textbf{60.1} & \textbf{63.9} & \textbf{43.5} \\
        \hdashedline
        \Block[c]{3-1}{\texttt{Qwen1.5}                                                                                                                                                                                                                                                                                                                                                                            \\[-4pt]\texttt{\footnotesize{(14B)}}}
                                                                                                                            & \texttt{ZSP} & 41.6                              & 17.4          & 28.1          & 53.3                             & 38.9          & 60.7          & 28.5                           & 42.0          & 33.8          & 20.5          & 35.3          & 37.3          & 25.3          \\
                                                                                                                            & \texttt{FSP} & 45.9                              & 25.4          & 31.6          & 61.4                             & 49.1          & 66.5          & 35.0                           & 47.6          & 43.4          & 27.9          & 38.6          & 38.7          & 29.2          \\
                                                                                                                            & \texttt{OUR} & \textbf{56.9}                     & \textbf{48.6} & \textbf{57.6} & \textbf{84.1}                    & \textbf{73.2} & \textbf{87.4} & \textbf{46.0}                  & \textbf{59.9} & \textbf{44.6} & \textbf{28.3} & \textbf{52.9} & \textbf{55.8} & \textbf{36.4} \\
        \hdashedline
        \Block[c]{3-1}{\texttt{InternLM2}                                                                                                                                                                                                                                                                                                                                                                          \\[-4pt]\texttt{\footnotesize{(20B)}}}
                                                                                                                            & \texttt{ZSP} & 42.3                              & 20.9          & 29.7          & 47.7                             & 31.9          & 55.9          & 29.8                           & 42.6          & 34.3          & 21.2          & 40.0          & 34.7          & 27.2          \\
                                                                                                                            & \texttt{FSP} & 55.9                              & 27.7          & 32.9          & 45.9                             & 38.2          & 65.3          & 31.3                           & 46.7          & 37.1          & 25.4          & 43.4          & 37.9          & 29.3          \\
                                                                                                                            & \texttt{OUR} & \textbf{57.8}                     & \textbf{53.1} & \textbf{60.5} & \textbf{83.9}                    & \textbf{72.3} & \textbf{91.1} & \textbf{49.7}                  & \textbf{59.0} & \textbf{48.2} & \textbf{31.8} & \textbf{55.9} & \textbf{63.3} & \textbf{40.5} \\
        \bottomrule
    \end{NiceTabular}
    \caption{
        The detailed \textbf{sentence} level correction $F_1$ score.
    }
    \label{tab:detail_results:scf}
\end{table*}

\begin{table*}[tb!]
    \centering
    \setlength{\tabcolsep}{4.25pt}
    \begin{NiceTabular}{lcccc|ccc|ccccccc}
        \toprule
        \rowcolor[gray]{.9}
        \textbf{Datasets}                                                                                                   &              & \Block[c]{1-3}{\textbf{rSighans}} &               &               & \Block[c]{1-3}{\textbf{ECSpell}} &               &               & \Block[c]{1-7}{\textbf{Lemon}} &               &               &               &               &               &               \\
        \textbf{Subsets}                                                                                                    &              & \textit{Y13}                      & \textit{Y14}  & \textit{Y15}  & \textit{Law}                     & \textit{Med}  & \textit{Odw}  & \textit{Car}                   & \textit{Cot}  & \textit{Enc}  & \textit{Gam}  & \textit{Med}  & \textit{New}  & \textit{Nov}  \\
        \midrule
        \rowcolor[gray]{.95}
        \Block[l]{1-15}{\texttt{Domain-Specific\;SOTAs} (\textit{Trained on in-domain gold-standard data of each dataset})} &              &                                   &               &               &                                  &               &               &                                &               &               &               &               &               &               \\
        \Block[l]{1-2}{\texttt{ReaLiSe}}                                                                                    &              & 85.0                              & 76.3          & 80.9          & \ood{48.7}                       & \ood{34.4}    & \ood{53.0}    & \ood{37.4}                     & \ood{42.7}    & \ood{32.9}    & \ood{16.3}    & \ood{33.8}    & \ood{35.1}    & \ood{23.2}    \\
        \midrule
        \rowcolor[gray]{.95}
        \Block[l]{1-15}{\texttt{Domain-General\;SOTAs} (\textit{Trained on about 34M synthetic CSC data})}                  &              &                                   &               &               &                                  &               &               &                                &               &               &               &               &               &               \\
        \Block[l]{1-2}{\texttt{Finetuned\;BERT}}                                                                            &              & 64.3                              & 51.0          & 57.2          & 66.3                             & 59.0          & 69.5          & 53.0                           & 64.1          & 46.0          & 35.6          & 52.3          & 57.5          & 36.3          \\
        \Block[l]{1-2}{\texttt{Softmasked\;BERT}}                                                                           &              & \textbf{65.6}                     & 49.3          & 57.3          & 67.2                             & 61.3          & 70.0          & 53.6                           & 63.3          & 45.4          & 31.6          & 51.0          & 57.9          & \textbf{38.5} \\
        \Block[l]{1-2}{\texttt{ReLM}}                                                                                       &              & 58.6                              & \textbf{51.1} & \textbf{61.0} & \textbf{68.3}                    & \textbf{63.9} & \textbf{73.0} & \textbf{54.4}                  & \textbf{66.1} & \textbf{48.2} & \textbf{37.5} & \textbf{55.1} & \textbf{60.5} & 37.1          \\
        \midrule
        \rowcolor[gray]{.95}
        \Block[l]{1-15}{\texttt{LLMs} (\textit{without CSC-specific training})}                                             &              &                                   &               &               &                                  &               &               &                                &               &               &               &               &               &               \\
        \Block[c]{3-1}{\texttt{Baichuan2}                                                                                                                                                                                                                                                                                                                                                                          \\[-4pt]\texttt{\footnotesize{(13B)}}}
                                                                                                                            & \texttt{ZSP} & 29.6                              & 11.2          & 14.5          & 20.5                             & 16.6          & 29.8          & 7.8                            & 7.4           & 12.5          & 4.1           & 11.9          & 14.2          & 10.6          \\
                                                                                                                            & \texttt{FSP} & 51.8                              & 29.7          & 34.0          & 54.9                             & 52.5          & 51.8          & 14.0                           & 35.3          & 23.0          & 9.5           & 29.5          & 39.0          & 26.2          \\
                                                                                                                            & \texttt{OUR} & \textbf{79.1}                     & \textbf{66.3} & \textbf{67.3} & \textbf{88.8}                    & \textbf{86.7} & \textbf{93.8} & \textbf{57.5}                  & \textbf{64.0} & \textbf{56.5} & \textbf{39.6} & \textbf{61.7} & \textbf{66.2} & \textbf{47.9} \\
        \hdashedline
        \Block[c]{3-1}{\texttt{Qwen1.5}                                                                                                                                                                                                                                                                                                                                                                            \\[-4pt]\texttt{\footnotesize{(14B)}}}
                                                                                                                            & \texttt{ZSP} & 48.8                              & 18.9          & 26.5          & 53.5                             & 35.4          & 58.1          & 27.1                           & 26.8          & 32.0          & 12.7          & 32.1          & 35.1          & 21.5          \\
                                                                                                                            & \texttt{FSP} & 51.0                              & 29.5          & 33.2          & 63.3                             & 44.4          & 66.9          & 22.7                           & 39.8          & 34.7          & 14.3          & 34.9          & 36.5          & 28.4          \\
                                                                                                                            & \texttt{OUR} & \textbf{75.2}                     & \textbf{62.8} & \textbf{66.0} & \textbf{88.6}                    & \textbf{84.5} & \textbf{91.6} & \textbf{52.4}                  & \textbf{62.9} & \textbf{49.6} & \textbf{34.3} & \textbf{54.6} & \textbf{59.5} & \textbf{42.6} \\
        \hdashedline
        \Block[c]{3-1}{\texttt{InternLM2}                                                                                                                                                                                                                                                                                                                                                                          \\[-4pt]\texttt{\footnotesize{(20B)}}}
                                                                                                                            & \texttt{ZSP} & 46.0                              & 18.1          & 27.3          & 40.5                             & 22.8          & 49.3          & 24.7                           & 31.9          & 29.7          & 12.3          & 31.0          & 29.2          & 26.6          \\
                                                                                                                            & \texttt{FSP} & 46.8                              & 25.5          & 33.4          & 56.7                             & 40.0          & 66.3          & 24.5                           & 34.2          & 30.4          & 10.4          & 40.9          & 32.9          & 28.9          \\
                                                                                                                            & \texttt{OUR} & \textbf{76.8}                     & \textbf{65.5} & \textbf{67.8} & \textbf{88.9}                    & \textbf{83.6} & \textbf{93.8} & \textbf{54.6}                  & \textbf{62.0} & \textbf{53.1} & \textbf{36.7} & \textbf{57.9} & \textbf{65.9} & \textbf{45.3} \\
        \bottomrule
    \end{NiceTabular}
    \caption{
        The detailed \textbf{character} level correction $F_1$ score.
    }
    \label{tab:detail_results:ccf}
\end{table*}

\begin{table*}[tb!]
    \centering
    \setlength{\tabcolsep}{4.25pt}
    \begin{NiceTabular}{lcccc|ccc|ccccccc}
        \toprule
        \rowcolor[gray]{.9}
        \textbf{Datasets}                                                                                                   &              & \Block[c]{1-3}{\textbf{rSighans}} &               &                 & \Block[c]{1-3}{\textbf{ECSpell}} &                 &                 & \Block[c]{1-7}{\textbf{Lemon}} &                 &                 &               &                 &                 &                 \\
        \textbf{Subsets}                                                                                                    &              & \textit{Y13}                      & \textit{Y14}  & \textit{Y15}    & \textit{Law}                     & \textit{Med}    & \textit{Odw}    & \textit{Car}                   & \textit{Cot}    & \textit{Enc}    & \textit{Gam}  & \textit{Med}    & \textit{New}    & \textit{Nov}    \\
        \midrule
        \rowcolor[gray]{.95}
        \Block[l]{1-15}{\texttt{Domain-Specific\;SOTAs} (\textit{Trained on in-domain gold-standard data of each dataset})} &              &                                   &               &                 &                                  &                 &                 &                                &                 &                 &               &                 &                 &                 \\
        \Block[l]{1-2}{\texttt{ReaLiSe}}                                                                                    &              & 13.0                              & 9.6           & 7.7             & \ood{10.6}                       & \ood{18.6}      & \ood{11.8}      & \ood{20.9}                     & \ood{13.4}      & \ood{20.8}      & \ood{22.5}    & \ood{16.5}      & \ood{16.7}      & \ood{22.6}      \\
        \Block[l]{1-2}{\citet{liu-etal-2023-chinese}}                                                                       &              & --                                & --            & --              & \wz7.4                           & \wz6.5          & \wz2.2          & --                             & --              & --              & --            & --              & --              & --              \\
        \midrule
        \rowcolor[gray]{.95}
        \Block[l]{1-15}{\texttt{Domain-General\;SOTAs} (\textit{Trained on about 34M synthetic CSC data})}                  &              &                                   &               &                 &                                  &                 &                 &                                &                 &                 &               &                 &                 &                 \\
        \Block[l]{1-2}{\texttt{Finetuned\;BERT}}                                                                            &              & 21.7                              & 16.5          & 12.5            & \textbf{\wz4.9}                  & 11.3            & \textbf{\wz2.9} & 12.3                           & \wz8.3          & 13.9            & 22.5          & \wz8.3          & \wz9.4          & 17.3            \\
        \Block[l]{1-2}{\texttt{Softmasked\;BERT}}                                                                           &              & 13.0                              & 17.6          & 14.5            & \wz6.1                           & 11.7            & \wz5.0          & 12.4                           & \wz7.1          & 14.8            & \textbf{20.4} & \wz9.6          & 10.6            & \textbf{16.6}   \\
        \Block[l]{1-2}{\texttt{ReLM}}                                                                                       &              & \textbf{\wz4.4}                   & \textbf{15.0} & \textbf{\wz9.5} & \wz7.8                           & \textbf{11.0}   & \wz7.1          & \textbf{12.1}                  & \textbf{\wz5.6} & \textbf{12.6}   & 20.8          & \textbf{\wz5.7} & \textbf{\wz8.4} & 17.5            \\
        \midrule
        \rowcolor[gray]{.95}
        \Block[l]{1-15}{\texttt{LLMs} (\textit{without CSC-specific training})}                                             &              &                                   &               &                 &                                  &                 &                 &                                &                 &                 &               &                 &                 &                 \\
        \Block[c]{3-1}{\texttt{Baichuan2}                                                                                                                                                                                                                                                                                                                                                                                          \\[-4pt]\texttt{\footnotesize{(13B)}}}
                                                                                                                            & \texttt{ZSP} & 34.8                              & 58.3          & 54.4            & 26.9                             & 43.1            & 21.0            & 40.6                           & 54.2            & 35.9            & 41.6          & 35.4            & 41.1            & 37.6            \\
                                                                                                                            & \texttt{FSP} & 21.7                              & 19.4          & 23.2            & \wz7.8                           & \textbf{\wz9.1} & \textbf{\wz0.4} & \wz8.3                         & \wz7.4          & 10.2            & 20.0          & \wz4.6          & \wz8.3          & \textbf{\wz7.7} \\
                                                                                                                            & \texttt{OUR} & \wz\textbf{8.7}                   & \textbf{14.1} & \wz\textbf{8.3} & \wz\textbf{4.5}                  & \wz\uline{9.9}  & \wz\textbf{0.4} & \wz\textbf{5.9}                & \wz\textbf{6.9} & \wz\textbf{8.9} & \textbf{19.2} & \wz\textbf{3.9} & \wz\textbf{5.7} & \uline{13.0}    \\
        \hdashedline
        \Block[c]{3-1}{\texttt{Qwen1.5}                                                                                                                                                                                                                                                                                                                                                                                            \\[-4pt]\texttt{\footnotesize{(14B)}}}
                                                                                                                            & \texttt{ZSP} & 34.8                              & 54.4          & 34.2            & \wz5.7                           & 35.4            & \wz2.1          & 18.5                           & 15.8            & 13.5            & 18.4          & 11.8            & 14.0            & 20.7            \\
                                                                                                                            & \texttt{FSP} & \textbf{15.9}                     & 30.9          & 31.7            & \wz5.3                           & \textbf{11.6}   & \textbf{\wz0.8} & \textbf{\wz8.9}                & 12.7            & \textbf{10.1}   & \textbf{14.7} & \wz9.5          & \wz\textbf{7.8} & \wz\textbf{5.5} \\
                                                                                                                            & \texttt{OUR} & \uline{21.7}                      & \textbf{19.6} & \textbf{10.2}   & \wz\textbf{4.9}                  & \uline{11.7}    & \wz\uline{2.9}  & \uline{11.2}                   & \wz\textbf{6.3} & \uline{14.8}    & \uline{29.4}  & \wz\textbf{5.4} & \uline{10.1}    & \uline{21.2}    \\
        \hdashedline
        \Block[c]{3-1}{\texttt{InternLM2}                                                                                                                                                                                                                                                                                                                                                                                          \\[-4pt]\texttt{\footnotesize{(20B)}}}
                                                                                                                            & \texttt{ZSP} & 65.2                              & 58.0          & 48.8            & 26.5                             & 50.7            & 17.7            & 28.8                           & 23.7            & 30.0            & 30.6          & 23.0            & 34.0            & 24.2            \\
                                                                                                                            & \texttt{FSP} & 21.7                              & 39.8          & 33.6            & 13.9                             & 30.7            & \wz2.5          & 18.2                           & 12.3            & 18.1            & 23.7          & 10.0            & 22.4            & 16.1            \\
                                                                                                                            & \texttt{OUR} & \textbf{13.0}                     & \textbf{16.5} & \wz\textbf{8.3} & \wz\textbf{2.5}                  & \textbf{12.4}   & \wz\textbf{0.4} & \wz\textbf{8.5}                & \wz\textbf{6.9} & \textbf{12.2}   & \textbf{22.5} & \wz\textbf{3.7} & \wz\textbf{6.1} & \textbf{15.1}   \\
        \bottomrule
    \end{NiceTabular}
    \caption{
        The detailed sentence level false positive rate.
    }
    \label{tab:detail_results:fpr}
\end{table*}

\subsection{Pre- \& Post-processing for Baselines}
\label{app:pre_post_processing}
In this study, we employ several pre- and post-processing techniques to mitigate the errors introduced by the limitations of baseline systems.
This ensures a fair comparison between our approach and the baselines.

\paragraph{BERT-based baselines}
Most current CSC models utilize BERT as the backbone.
However, BERT presents challenges that can degrade performance during evaluation:
\begin{inparaenum}[1)]
    \item \textit{Full-width Punctuation:} BERT's tokenization process may normalize full-width punctuation to half-width, leading to numerous unnecessary punctuation replacements. To counter this, we prevent the model from modifying the original punctuation;
    \item \textit{Special Tokens:} BERT-based models may predict a special `\texttt{[UNK]}` token in some cases, resulting in the removal of the original character. In these instances, we retain the original character when a special token is predicted;
    \item \textit{Input Length Limitation:} BERT-based models show limited generalization beyond their maximum training length. We truncate inputs to a maximum length of 128 characters and concatenate the remaining characters to the output.
\end{inparaenum}

\paragraph{LLM baselines}
The outputs of LLMs sometimes fail to align with evaluation, primarily due to their inadequate instruction-following capability.
To address this, we apply specific rules for post-processing:
\begin{inparaenum}[1)]
    \item \textit{Redundant Phrases:} We remove redundant phrases such as ``修改后的句子是：'' (\textit{The corrected sentence is:}), identified through common patterns input in the model output;
    \item \textit{Redundant Punctuation:} Many sentences in the dataset lack terminal periods, yet some models inappropriately add them. To prevent incorrect evaluations due to this discrepancy, we remove any added terminal period if the original sentence did not have one.
\end{inparaenum}

\section{Details of Evaluation}
\label{app:evaluation_metrics}
\subsection{Evaluation Metrics}

In this work, we use the following metrics to evaluate the performance of our approach and the baselines.
\paragraph{Sentence-level Correction $F_1$ (S-F)}
S-F consists of two parts: precision (S-P) and recall (S-R):
\begin{equation}
    \text{S-F} = 2 \times \frac{\text{S-P} \times \text{S-R}}{\text{S-P} + \text{S-R}}.
\end{equation}
where $\text{S-P}$ represents the proportion of correctly corrected sentences among all sentences modified by the model, and $\text{S-R}$ represents the proportion of correctly corrected sentences among all sentences need to be corrected.

A sentence is considered correctly corrected if and only if \textbf{all} errors in the sentence are fixed and no new errors are introduced.
This strict definition makes the sentence-level $F_1$ score rigorous, but also makes it vulnerable when the number of evaluation samples is limited, such as ECSpell dataset, which contains only 500 sentences for each sub-domain, and lacks of ability to detect subtle differences between models when evaluating on the same dataset, which a sentence contains multiple errors.

\paragraph{Character-level Correction $F_1$ (C-F)}
Different from the sentence-level $F_1$ score, the character-level $F_1$ score focuses on the correctness of each character in the sentence.
Similar to the sentence-level $F_1$ score, the character-level $F_1$ score also consists of two parts: precision (C-P) and recall (C-R).
C-P is the proportion of correctly corrected characters among all characters modified by the model, and C-R is the proportion of correctly corrected characters among all characters need to be corrected.

Conventional character-level metrics of CSC are based on point-wise evaluation, which fall short when models insert or delete characters, as they can inaccurately mark all subsequent characters as incorrect due to a single addition or deletion.
To overcome this, \textbf{we implement Levenshtein algorithm to align the model output with the target sentence} and calculate the character-level metrics based on the aligned results.
This alignment-based method provides a more reasonable evaluation of character-level performance.

\paragraph{Sentence-level False Positive Rate (FPR)}
Both sentence-level $F_1$ score and character-level $F_1$ score overlook the cases where the model introduces unnecessary modifications to a de-facto correct sentence.
To fill this gap, sentence-level False Positive Rate (FPR) is proposed to measure the proportion of sentences that are initially correct but modified by the model.

\subsection{Evaluation Settings and Conventions}
During evaluation, we \textbf{remove all whitespaces} and \textbf{convert all full-width punctuation to half-width} from the input and output sentences to guarantee a fair comparison\rlap{.}\footnote{BERT-based models often remove whitespaces during tokenization and may convert full-width punctuation to half-width when correcting spelling errors (e.g., \texttt{ReLM}).}

When evaluating the Lemon dataset, we \textbf{ignore} all sentences where the \textbf{input and output sentence lengths do not match}, \textit{following the dataset's convention}.

\section{More Results}
\subsection{Detailed Results}
\label{app:subdomain_results}
Due to the space limitation, we only present the average results of each dataset in the main text.
The detailed results of each dataset are shown in Table~\ref{tab:detail_results:scf}, Table~\ref{tab:detail_results:ccf}, and Table~\ref{tab:detail_results:fpr}.

\begin{table}[t!]
    \setlength{\tabcolsep}{2.5pt}
    \renewcommand{\arraystretch}{1.1}
    \centering
    {
        \scalebox{0.9}{
            \begin{NiceTabular}{ll}
                \toprule
                \rowcolor[gray]{.95}
                Input                 & 商务部\wrong{前}头，$11$月底完成                                       \\
                Reference             & 商务部\correct{牵}头，$11$月底完成                                     \\
                \midrule
                \texttt{ReLM}         & \textcolor{gray}{商务部}\correct{牵}头\textcolor{gray}{，$11$月底完成} \\
                \hdashedline
                \texttt{BC2\;13B ZSP} & \textcolor{gray}{商务部}\wrong{前面}\textcolor{gray}{，$11$月底完成}   \\
                \texttt{BC2\;13B FSP} & \textcolor{gray}{商务部}\wrong{日前}\textcolor{gray}{，$11$月底完成}   \\
                \hdashedline
                \texttt{BC2\;13B OUR} & \textcolor{gray}{商务部}\correct{牵}头\textcolor{gray}{，$11$月底完成} \\
                \midrule
                \midrule
                \rowcolor[gray]{.95}
                Input                 & \wrong{虎}珀酸索\wrong{莉}那新片主要功能是什么                              \\
                Reference             & \correct{琥}珀酸索\correct{利}那新片主要功能是什么                          \\
                \midrule
                \texttt{ReLM}         & \correct{琥}珀酸索\wrong{莉}那新片\textcolor{gray}{主要功能是什么}          \\
                \hdashedline
                \texttt{BC2\;13B ZSP} & \correct{琥}珀酸索\correct{利}那新片\textcolor{gray}{主要功能是什么}        \\
                \texttt{BC2\;13B FSP} & \wrong{虎}珀酸索\wrong{莉}那新片\textcolor{gray}{主要功能是什么}            \\
                \hdashedline
                \texttt{BC2\;13B OUR} & \correct{琥}珀酸索\correct{利}那新片\textcolor{gray}{主要功能是什么}        \\
                \bottomrule
            \end{NiceTabular}
        }
    }
    \caption{
        Qualitative examples of our approach and the baselines.
        Corrections marked in ``\correct{Blue}'' are correct, while those in ``\wrong{Red}'' are incorrect.
    }
    \label{tab:qualitative_examples}
\end{table}

\subsection{Qualitative Examples}
\label{app:qualitative_examples}
We provide two qualitative examples to illustrate the performance of our approach in Table~\ref{tab:qualitative_examples}.

In the first case (``\textit{Led by the Ministry of Commerce, to be completed by the end of November}''), the word ``牵头'' (\textit{qiāntóu, led by}) is misspelled as ``\wrong{前}头'' (\textit{qiántóu, front}) in the input sentence.
Both the ZSP and FSP baselines mistakenly put their attention on the character ``前'' (\textit{front}) and incorrectly correct ``前头'' to ``日前'' (\textit{a few days ago}) and ``前面'' (\textit{front}), respectively.
Such corrections are not only implausible but also linguistically awkward.
In contrast, the domain-general model ReLM and our approach successfully correct the misspelling.

In the second case (``\textit{What are the main functions of Solifenacin Succinate Tablets}''), the name of the drug ``琥珀酸索利那新片'' (\textit{Solifenacin Succinate Tablets}) is misspelled.
To correct the misspelling, the knowledge of the medical domain is required.
In this case, the ReLM model fails to correct the misspelling, while the zero-shot prompting baseline and our approach successfully correct it.
It is worth noting that the few-shot prompting baseline also fails to correct the misspelling, which indicates that the inclusion of inappropriate examples may lead to worse performance.

\section{More Discussions}
\label{app:more_discussions}
\subsection{Impact of the Pre-training Data}
\label{app:pre_training_data}
\begin{table*}[tb!]
    \centering
    \begin{NiceTabular}{lccc>{\columncolor{figure_light_red!6}}c|cc>{\columncolor{figure_light_red!6}}c|cc>{\columncolor{figure_light_red!6}}c}
        \toprule
        \rowcolor[gray]{1.0}
        \Block[l]{2-1}{\textbf{System}} & \Block[c]{2-1}{\textbf{Data}                                                                                                             \\[-3pt]\textbf{Amount}}& \Block[c]{1-3}{\textbf{rSighan} \textit{15}} &           &           & \Block[c]{1-3}{\textbf{Lemon} \textit{Nov}} &           &           & \Block[c]{1-3}{\textbf{ECSpell} \textit{Odw}} &           &           \\
        \rowcolor[gray]{1.0}
                                        &                              & S-F\bgood & C-F\bgood & FPR\sgood & S-F\bgood & C-F\bgood & FPR\sgood & S-F\bgood & C-F\bgood & FPR\sgood \\
        \midrule
        \texttt{GPT2\ \ \ \ 1.5B}       & Small                        & 56.6      & 64.4      & 10.4      & 26.1      & 31.8      & 31.4      & 82.8      & 85.8      & 5.5       \\
        \hdashedline
        \texttt{Qwen1.5\ 463M}          & \Block[c]{2-1}{Large}        & 56.3      & 63.5      & 10.0      & 33.2      & 40.2      & 22.2      & 84.7      & 89.9      & 3.8       \\
        \texttt{Qwen1.5\ 1.8B}          &                              & 58.3      & 65.3      & 10.3      & 35.6      & 42.3      & 19.9      & 90.3      & 92.8      & 1.7       \\
        \bottomrule
    \end{NiceTabular}
    \caption{
        A brief comparison of the performance of LLMs of different sizes and pre-training data amounts on three datasets.
    }
    \label{tab:ablation:lm_size_vs_data}
\end{table*}

There are two main factors that differentiate LLMs from simpler LMs: the scale of pre-training data and the model size.
The impact of model size on the performance of LLMs has been discussed in \S\ref{sec:ablation:size}.
In this subsection, we aim to investigate the impact of pre-training data on the performance of our approach.

We compare \texttt{Qwen1.5}, a recent LLM family, with \texttt{GPT2}, which also has a causal LM (decoder-only) architecture.
The \texttt{GPT2} model family partially overlaps in model size with the \texttt{Qwen1.5} model family, but it was trained on a much smaller dataset, CLUECorpusSmall.
The CLUECorpusSmall dataset contains only about 5 billion characters and has limited diversity in text sources, including only news, Wikipedia, forums, and comments.

As shown in Table~\ref{tab:ablation:lm_size_vs_data}, when the model sizes are similar, the \texttt{Qwen1.5} model family outperforms the \texttt{GPT2} model family on all three datasets.
The largest performance gap is observed on the Lemon-Nov dataset, where a smaller 463M \texttt{Qwen1.5} model even outperforms a larger 1.5B \texttt{GPT2} model by 7.1\% in the sentence-level correction $F_1$ score.
This is because the Lemon-Nov dataset contains texts from the novel domain, which is not included in the CLUECorpusSmall dataset.
These results indicate that the scale and diversity of the pre-training data are crucial for the performance of our approach.

\begin{table}[tb!]
    \centering
    \setlength{\tabcolsep}{3.8pt}
    \scalebox{0.85}{
        \begin{NiceTabular}{lcccc|c}
            \toprule
            \rowcolor[gray]{1.0}
            \Block[l]{2-1}{\textbf{System}}        & \Block[c]{2-1}{\textbf{Method}} & \Block[c]{1-3}{\textbf{ECSpell}} &              &              & \textbf{CSCD-NS} \\
                                                   &                                 & \textit{Law}                     & \textit{Med} & \textit{Odw} & \textit{Test}    \\
            \midrule
            \Block[l]{2-1}{\texttt{BERT-based}}    & \texttt{34M-ft}                 & 60.4                             & 50.9         & 66.5         & 51.0             \\
                                                   & \texttt{Id-ft}                  & 91.2                             & 82.4         & 83.6         & 66.2 (73.6)      \\
            \hdashedline
            \Block[l]{2-1}{\texttt{GPT2\,110M}}    & \texttt{Id-ft}                  & 71.2                             & 35.6         & 53.8         & --               \\
                                                   & \texttt{OUR}                    & 66.4                             & 60.0         & 78.6         & --               \\
            \hdashedline
            \Block[l]{2-1}{\texttt{Baichuan2\,7B}} & \texttt{Id-ft}                  & 86.0                             & 73.2         & 82.6         & 56.4 (64.4)      \\
                                                   & \texttt{OUR}                    & 82.1                             & 79.7         & 89.7         & 62.7             \\
            \bottomrule
        \end{NiceTabular}
    }
    \caption{
        The \textbf{S-F} of models supervised fine-tuned and those from our approach.
        \texttt{Id-ft} denotes the model fine-tuned on the in-domain training data of either ECSpell or CSCD-NS.
        Scores in parentheses represent the \textbf{S-F} of the model, which was pre-trained on 2M carefully crafted synthetic CSC data prior to being fine-tuned on the in-domain training data.
    }
    \label{tab:discussion:sft}
\end{table}

\subsection{Comparison to the Supervised Fine-tuning Method}
\label{app:sft_comparison}
In this subsection, we compare our approach with the supervised fine-tuning method.

However, we did not fine-tune the LLMs ourselves, as fine-tuning an LLM on the 34M synthetic CSC data would be extremely time-consuming and computationally expensive.
Additionally, the supervised fine-tuning method typically requires careful hyperparameter tuning to achieve the best performance, further increasing the computational cost.

Instead, we leverage the findings from \citet{li-etal-2023-ineffectiveness}, who fine-tuned the \texttt{Baichuan2\,7B} and \texttt{GPT2} models on the ECSpell dataset, and \citet{hu-etal-2024-cscd}, who fine-tuned the \texttt{Baichuan2\,7B} model on the CSCD-NS dataset.

The results are shown in Table~\ref{tab:discussion:sft}.
Compared to the BERT-based models, the supervised fine-tuning method is less effective in improving the performance of causal LMs like \texttt{GPT2} and recent LLMs such as \texttt{Baichuan2}.

Our training-free approach even outperforms the supervised fine-tuning counterpart on the Med and Odw sub-domains of the ECSpell dataset.
This phenomenon can be attributed to the characteristics of the ECSpell dataset, which, as pointed out by \citet{wu-etal-2023-rethinking}, contains a high proportion (more than 70\%) of error-correction pairs that never appeared in the training data.
The supervised fine-tuning method is not effective in handling these unseen error-correction pairs, whereas our approach can still correct them.

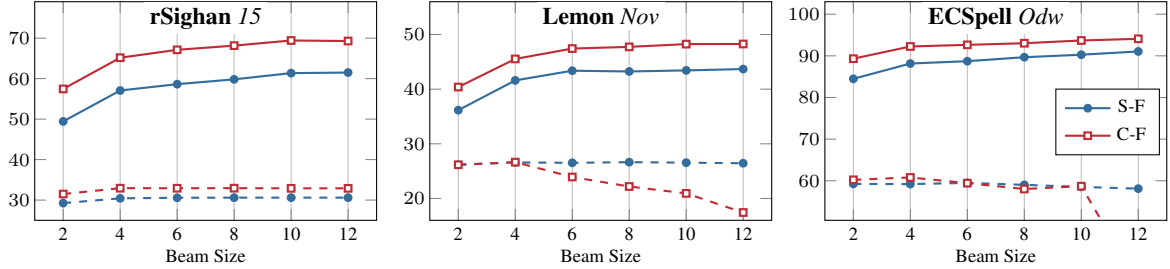
\begin{figure*}[tb!]
    \centering
    \captionsetup[subfigure]{skip=-1pt, margin=5pt}
    \begin{tikzpicture}[
            legend/.style={
                    fill=white,
                    font=\footnotesize,
                    inner sep=2pt,
                    minimum width=0.8cm,
                    text opacity=1.0,
                    fill opacity=1.0,
                },
            diff label/.style={
                    font=\scriptsize,
                    inner sep=0.5pt,
                    outer sep=1.5pt,
                    fill=white,
                    fill opacity=0.9,
                    text opacity=1.0,
                    rounded corners=1pt,
                },
            trim left
        ]
        \centering
        \begin{groupplot}[
                group style={
                        group size=3 by 1,
                        x descriptions at=edge bottom,
                        horizontal sep=0.7cm,
                        vertical sep=0.1cm,
                    },
                width=0.38\linewidth,
                height=0.28\linewidth,
                xlabel={Beam Size},
                xmin=1,
                xmax=13,
                xmajorgrids=true,
                ytick={0, 10, 20, 30, 40, 50, 60, 70, 80, 90, 100},
                xtick={2, 4, 6, 8, 10, 12},
                xticklabels={2, 4, 6, 8, 10, 12},
                /tikz/font=\scriptsize,
                ylabel shift=-4pt,
                xlabel shift=-4pt,
                yticklabel shift=-2pt,
                xticklabel shift=-1pt,
                legend style={
                        at={(0.98,0.45)},
                        anchor=east,
                    },
            ]
            \nextgroupplot[ymin=25,ymax=79]
            \addplot+ [mark=*, draw=figure_blue, thick,
                mark size=1.25pt,
                mark options={fill=figure_blue, fill opacity=1.0, solid},
                opacity=1.0,
            ] table [row sep=\\] {
                    x	y\\
                    2	49.432\\
                    4	57.065\\
                    6	58.633\\
                    8	59.823\\
                    10	61.362\\
                    12	61.498\\
                };
            \addplot+ [mark=square*, draw=figure_red, thick,
                mark size=1.25pt,
                mark options={fill=white, fill opacity=1.0, solid},
                opacity=1.0,
            ] table [row sep=\\] {
                    x	y	ey+	ey-\\
                    2	57.449\\
                    4	65.164\\
                    6	67.114\\
                    8	68.152\\
                    10	69.422\\
                    12	69.286\\
                };
            \addplot+ [mark=*, draw=figure_blue, thick, dashed,
                mark size=1.25pt,
                mark options={fill=figure_blue, fill opacity=1.0, solid},
                opacity=1.0,
            ] table [row sep=\\] {
                    x	y\\
                    2	29.248\\
                    4	30.431\\
                    6	30.574\\
                    8	30.6\\
                    10	30.6\\
                    12	30.6\\
                };
            \addplot+ [mark=square*, draw=figure_red, thick, dashed,
                mark size=1.25pt,
                mark options={fill=white, fill opacity=1.0, solid},
                opacity=1.0,
            ] table [row sep=\\] {
                    x	y	ey+	ey-\\
                    2	31.492\\
                    4	32.94\\
                    6	32.923\\
                    8	32.941\\
                    10	32.903\\
                    12	32.903\\
                };
            \nextgroupplot[ymin=16,ymax=56]
            \addplot+ [mark=*, draw=figure_blue, thick,
                mark size=1.25pt,
                mark options={fill=figure_blue, fill opacity=1.0, solid},
                opacity=1.0,
            ] table [row sep=\\] {
                    x	y\\
                    2	36.156\\
                    4	41.602\\
                    6	43.374\\
                    8	43.235\\
                    10	43.433\\
                    12	43.676\\
                };
            \addplot+ [mark=square*, draw=figure_red, thick,
                mark size=1.25pt,
                mark options={fill=white, fill opacity=1.0, solid},
                opacity=1.0,
            ] table [row sep=\\] {
                    x	y	ey+	ey-\\
                    2	40.384\\
                    4	45.53\\
                    6	47.425\\
                    8	47.74\\
                    10	48.242\\
                    12	48.27\\
                };
            \addplot+ [mark=*, draw=figure_blue, thick, dashed,
                mark size=1.25pt,
                mark options={fill=figure_blue, fill opacity=1.0, solid},
                opacity=1.0,
            ] table [row sep=\\] {
                    x	y\\
                    2	26.16\\
                    4	26.587\\
                    6	26.522\\
                    8	26.637\\
                    10	26.548\\
                    12	26.442\\
                };
            \addplot+ [mark=square*, draw=figure_red, thick, dashed,
                mark size=1.25pt,
                mark options={fill=white, fill opacity=1.0, solid},
                opacity=1.0,
            ] table [row sep=\\] {
                    x	y	ey+	ey-\\
                    2	26.164\\
                    4	26.645\\
                    6	23.922\\
                    8	22.188\\
                    10	20.924\\
                    12	17.427\\
                };
            \nextgroupplot[ymin=50.5,ymax=103]
            \addplot+ [mark=*, draw=figure_blue, thick,
                mark size=1.25pt,
                mark options={fill=figure_blue, fill opacity=1.0, solid},
                opacity=1.0,
            ] table [row sep=\\] {
                    x	y\\
                    2	84.48\\
                    4	88.155\\
                    6	88.716\\
                    8	89.668\\
                    10	90.272\\
                    12	91.051\\
                };
            \addplot+ [mark=square*, draw=figure_red, thick,
                mark size=1.25pt,
                mark options={fill=white, fill opacity=1.0, solid},
                opacity=1.0,
            ] table [row sep=\\] {
                    x	y	ey+	ey-\\
                    2	89.318\\
                    4	92.249\\
                    6	92.64\\
                    8	93.03\\
                    10	93.687\\
                    12	94.088\\
                };
            \addplot+ [mark=*, draw=figure_blue, thick, dashed,
                mark size=1.25pt,
                mark options={fill=figure_blue, fill opacity=1.0, solid},
                opacity=1.0,
            ] table [row sep=\\] {
                    x	y\\
                    2	59.244\\
                    4	59.197\\
                    6	59.494\\
                    8	59.017\\
                    10	58.526\\
                    12	58.106\\
                };
            \addplot+ [
                mark=square*, draw=figure_red, thick, dashed,
                mark size=1.25pt,
                mark options={fill=white, fill opacity=1.0, solid},
                opacity=1.0,
            ] table [row sep=\\] {
                    x	y	ey+	ey-\\
                    2	60.229\\
                    4	60.792\\
                    6	59.426\\
                    8	58.017\\
                    10	58.698\\
                    12	35.329\\
                };
            \legend{S-F, C-F};
        \end{groupplot}
        \node[anchor=north, legend] at (group c1r1.north) {\textbf{rSighan} \textit{15}};
        \node[anchor=north, legend] at (group c2r1.north) {\textbf{Lemon} \textit{Nov}};
        \node[anchor=north, legend] at (group c3r1.north) {\textbf{ECSpell} \textit{Odw}};
    \end{tikzpicture}
    \caption{
        The scores of \texttt{Baichuan2\;7B} with different beam sizes.
        The solid lines represent the results of our approach, and the dashed lines represent the results of the few-shot baseline.
        We can observe that larger beam sizes may lead to worse C-F scores in few-shot settings.
    }
    \label{fig:beam:baichuan2}
\end{figure*}

\subsection{Influence of Beam Size}
\label{app:beam_size}
During searching the most likely correction sequence, the beam search algorithm is used to avoid the exponential growth of the search space and the local minimum caused by greedy search.
Knowing the impact of the beam size on the performance helps researchers to choose a proper beam size to balance the trade-off between the performance and the computational cost.
The results are shown in Figure~\ref{fig:beam:baichuan2}.
Though the larger beam size consistently leads to better performance, the improvement becomes marginal when the beam size is larger than~6.

\begin{table}[tb!]
    \setlength{\tabcolsep}{4.0pt}
    \renewcommand{\arraystretch}{0.95}
    \centering
    \scalebox{0.90}{
        \begin{NiceTabular}{lcc|cc|cc}
            \toprule
                                                                              & \Block[c]{1-2}{\footnotesize{\textbf{rSighan} \textit{15}}} &                  & \Block[c]{1-2}{\footnotesize{\textbf{Lemon} \textit{Nov}}} &                  & \Block[c]{1-2}{\footnotesize{\textbf{ECSpell} \textit{Odw}}} &                 \\[-2pt]
                                                                              & \texttt{Dev}                                                & \texttt{True}    & \texttt{Dev}                                               & \texttt{True}    & \texttt{Dev}                                                 & \texttt{True}   \\
            \midrule
            \rowcolor[gray]{.95}
            \Block[l]{1-7}{\textit{Distortion Model:} $\log p_{\mathtt{DM}}$} &                                                             &                  &                                                            &                  &                                                              &                 \\
            \texttt{Idt.}                                                     & \ewm0.04                                                    & \ewm0.03         & \ewm0.04                                                   & \ewm0.02         & \ewm0.04                                                     & \ewm0.02        \\
            \texttt{Sa.P.}                                                    & \ewm3.75                                                    & \ewm4.00         & \ewm3.75                                                   & \ewm4.66         & \ewm3.75                                                     & \ewm4.17        \\
            \texttt{Si.P.}                                                    & \ewm4.85                                                    & \ewm5.02         & \ewm4.85                                                   & \ewm5.45         & \ewm4.85                                                     & \ewm5.87        \\
            \texttt{Si.S.}                                                    & \ewm5.40                                                    & \ewm8.63         & \ewm5.40                                                   & \ewm8.04         & \ewm5.40                                                     & \ewm6.66        \\
            \midrule
            \midrule
            S-F\bgood                                                         & \wm59.8                                                     & \textbf{\ewp0.9} & \wm43.2                                                    & \wm0.0           & \wm89.7                                                      & \uline{\ewm0.8} \\
            C-F\bgood                                                         & \wm68.2                                                     & \textbf{\ewp1.4} & \wm47.7                                                    & \textbf{\ewp0.2} & \wm93.0                                                      & \uline{\ewm0.3} \\
            \rowcolor{figure_light_red!6}
            FPR\sgood                                                         & \wm\wz8.1                                                   & \wz0.0           & \wm13.6                                                    & \uline{\ewp0.3}  & \wz\wm1.3                                                    & \wm0.0          \\
            \bottomrule
        \end{NiceTabular}
    }
    \caption{
        The impact of distortion model on the performance of \texttt{Baichuan2\;7B}.
        ``\texttt{True}'' denotes that the distortion model is derived from the \textbf{true} distortion distribution of each dataset.
        `` \texttt{Dev}'' represents the distortion model from the Pseudo-Dev.
    }
    \label{tab:analysis:distortion}
\end{table}

\subsection{Effectiveness of the Estimated Distortion Model}
\label{app:distortion_analysis}
The distortion model is a key component in our approach.
In this work, we utilize a minimal distortion model and directly estimate the distortion probabilities from the statistics of the Pseudo-Dev dataset.
Obviously, this estimation will be different from the true probabilities.

To verify the effectiveness of the estimated distortion model, we conduct experiments comparing the estimated distortion model with the true distortion model.
The results are presented in Table~\ref{tab:analysis:distortion}.
The upper part of the table shows the difference between the estimated distortion model and the true distortion model.
We can see that the estimated one is quite close to the true one, except for the \texttt{Similar\,Shape} distortion type.
The lower part shows that the difference between the performance is marginal, indicating that the estimated distortion model is sufficient for our approach to achieve a good performance, and has good generalization ability across different datasets.

\subsection{Inference Speed}
\label{app:inference_speed}

\begin{table}[tb!]
    \centering
    \scalebox{0.9}{
        \begin{NiceTabular}{lccc}
            \toprule
            \rowcolor[gray]{1.0}
            \Block[l]{2-2}{\textbf{System}}         &              & \Block[c,fill=gray!10]{1-2}{\textbf{Inference Speed} (ms)} &                             \\
            \rowcolor[gray]{1.0}
                                                    &              & \textit{per} \textbf{Sent.}                                & \textit{per} \textbf{Char.} \\
            \midrule
            \texttt{ReLM}                           &              & \phantom{0,0}14.4                                          & \wz0.4                      \\
            \midrule
            \Block[l]{3-1}{\texttt{Baichuan2\;13B}} & \texttt{ZSP} & \phantom{0,}899.8                                          & 22.2                        \\
                                                    & \texttt{FSP} & 1,057.4                                                    & 26.1                        \\
                                                    & \texttt{OUR} & 1,541.0                                                    & 38.0                        \\
            \bottomrule
        \end{NiceTabular}
    }
    \caption{
        The inference speed of different models.
    }
    \label{tab:runtime}
\end{table}

We conducted a brief runtime analysis to evaluate the inference speed of our approach.
The analysis was performed using a single NVIDIA A100 40GB GPU with an Intel Xeon Gold 6248R (3.00GHz) CPU.
The batch size was set to 1 for all models, and other hyperparameters were set to the same values as in the main experiments.

The average inference speed of each model on the ECSpell-Odw dataset is shown in Table~\ref{tab:runtime}.
Due to the large model size and the autoregressive decoding process, LLMs are significantly slower than the BERT-based ReLM model.
Compared to the ZSP and FSP baselines, our approach is slower (1.71$\times$ and 1.45$\times$, respectively), primarily due to our immature implementation of the distortion model, which can be further optimized to improve inference speed.

\begin{table}[tb!]
    \setlength{\tabcolsep}{4.5pt}
    \centering
    \scalebox{0.85}{
        \begin{NiceTabular}{lcc@{\!\,}cc>{\columncolor{figure_light_red!6}}c;>{\columncolor{figure_light_red!6}}cc}
            \toprule
            \Block[l]{1-2}{{System}}                             &                              & Ctx.      & S-F\bgood     & C-F\bgood     & FPR\sgood\phantom{i} & CER\sgood     & \!CERR\hspace{-0.1em}\bgood \\
            \midrule
            \Block[l]{1-2}{No correction}                        &                              &           & --            & --            & --                   & 4.83          & --                          \\
            \midrule
            \rowcolor[gray]{.95}
            \Block[l]{1-7}{\texttt{Domain-Specific\;SOTAs}}      &                              &           &               &               &                      &               &                             \\
            \Block[l]{1-2}{\citet{leng-etal-2021-fastcorrect}}   &                              & \ding{55} & --            & --            & --                   & 4.16          & 13.9                        \\
            \Block[l]{1-2}{\citet{leng-etal-2021-fastcorrect-2}} &                              & \ding{55} & --            & --            & --                   & 4.11          & 14.9                        \\
            \Block[l]{1-2}{\citet{leng-etal-2023-softcorrect}}   &                              & \ding{55} & --            & --            & --                   & 3.57          & 26.1                        \\
            \midrule
            \rowcolor[gray]{.95}
            \Block[l]{1-7}{\texttt{Domain-General\;SOTAs}}       &                              &           &               &               &                      &               &                             \\
            \Block[l]{2-2}{\texttt{Finetuned\;BERT}}             &                              & \ding{55} & \textbf{23.7} & \textbf{25.7} & 5.3                  & \textbf{4.39} & \wz\textbf{9.1}             \\
                                                                 &                              & \ding{51} & 18.2          & 19.5          & \textbf{1.8}         & 4.43          & \wz8.3                      \\
            \hdashedline
            \Block[l]{2-2}{\texttt{Softmasked\;BERT}}            &                              & \ding{55} & \textbf{22.6} & \textbf{25.5} & 5.4                  & 4.43          & \wz8.3                      \\
                                                                 &                              & \ding{51} & 19.8          & 21.4          & \textbf{1.9}         & \textbf{4.39} & \wz\textbf{9.1}             \\
            \hdashedline
            \Block[l]{2-2}{\texttt{ReLM}}                        &                              & \ding{55} & \textbf{24.7} & \textbf{27.5} & 4.7                  & \textbf{4.30} & \textbf{11.0}               \\
                                                                 &                              & \ding{51} & 17.7          & 18.6          & \textbf{2.5}         & 4.50          & \wz6.8                      \\
            \midrule
            \rowcolor[gray]{.95}
            \Block[l]{1-7}{\texttt{LLMs}}                        &                              &           &               &               &                      &               &                             \\
            \Block[c]{2-1}{\texttt{Baichuan2}                                                                                                                                                                    \\[-4pt]\texttt{\footnotesize{(13B)}}}
                                                                 & \Block[c]{2-1}{\texttt{OUR}} & \ding{55} & 34.8          & 43.1          & 3.8                  & 3.68          & 23.8                        \\
                                                                 &                              & \ding{51} & \textbf{41.7} & \textbf{51.3} & \textbf{3.0}         & \textbf{3.29} & \textbf{31.9}               \\
            \hdashedline
            \Block[c]{2-1}{\texttt{Qwen1.5}                                                                                                                                                                      \\[-4pt]\texttt{\footnotesize{(14B)}}}
                                                                 & \Block[c]{2-1}{\texttt{OUR}} & \ding{55} & 28.7          & 37.4          & 7.1                  & 4.10          & 15.1                        \\
                                                                 &                              & \ding{51} & \textbf{38.0} & \textbf{48.5} & \textbf{4.4}         & \textbf{3.44} & \textbf{28.8}               \\
            \hdashedline
            \Block[c]{2-1}{\texttt{InternLM2}                                                                                                                                                                    \\[-4pt]\texttt{\footnotesize{(20B)}}}
                                                                 & \Block[c]{2-1}{\texttt{OUR}} & \ding{55} & 33.8          & 42.6          & 4.1                  & 3.70          & 23.4                        \\
                                                                 &                              & \ding{51} & \textbf{40.4} & \textbf{51.3} & \textbf{3.0}         & \textbf{3.29} & \textbf{31.9}               \\
            \bottomrule
        \end{NiceTabular}
    }
    \caption{
        Results of contextual enhanced spelling correction on {AISHELL-1} dataset.
        \ding{51} denotes the results of models taking 3 preceding sentences as the input prefix.
        All the preceding context are also predicted by the same ASR model.
    }
    \label{tab:asr:cxt_results}
\end{table}

\subsection{Context as New Knowledge}
\label{app:context_as_new_knowledge}

In Section~\ref{app:new_knowledge}, we used a toy example to demonstrate that our approach can introduce new knowledge into the LLM by merely modifying the input prefix.
However, in real-world scenarios, it is difficult to automatically extract the key characters as we did in the toy example and ensure they are suitable for the input prefix.
Luckily, sentences in real-world contexts are not isolated but are part of a paragraph, and their preceding sentences can provide valuable information for error correction.
Thus, we can treat the preceding context as new knowledge and introduce it into the LLM.

Since existing datasets for CSC are composed of isolated sentences, it is impossible to validate the effectiveness of using the preceding context as new knowledge on them.
Therefore, we utilize the ASR error correction dataset derived from \textbf{AISHELL-1} \cite{bu-etal-2017-aishell}, where the sentences are consecutive and part of coherent passages.
In this dataset, \cite{leng-etal-2021-fastcorrect} used an ASR model to transcribe the speech data, introducing spelling errors naturally caused by the ASR system.

In addition to conventional CSC metrics, we also report the \textbf{Character Error Rate} (\textbf{CER})\footnote{CER calculates the number of insertions, deletions, and substitutions edits required to transform the predicted sequence into the target sequence:\begin{align*} \mathrm{CER} = \frac{n_{\texttt{insert}} + n_{\texttt{delete}} + n_{\texttt{replace}}}{n_{\texttt{target}}}. \end{align*}} and \textbf{Character Error Rate Reduction} (\textbf{CERR})\footnote{CERR represents the percentage of CER reduction compared to the baseline model:\begin{align*} \mathrm{CERR} = 1 - \frac{\mathrm{CER}_{\text{ours}}}{\mathrm{CER}_{\text{baseline}}}. \end{align*}} to compare with the baseline models \cite{leng-etal-2021-fastcorrect,leng-etal-2021-fastcorrect-2,leng-etal-2023-softcorrect}.

Specifically, we take the three preceding sentences from the source side as the new knowledge:
\begin{equation}
    \boldsymbol{k} = \boldsymbol{x}_{-3} \oplus \boldsymbol{x}_{-2} \oplus \boldsymbol{x}_{-1}.
\end{equation}
where $\boldsymbol{x}_{-i}$ denotes the $i$-th sentence preceding the current one.

The results in Table~\ref{tab:asr:cxt_results} clearly show that our method can effectively utilize the preceding context as new knowledge to improve the performance of ASR error correction.
Meanwhile, we observe that the BERT-based baselines cannot effectively utilize the preceding context to achieve better performance.

\end{CJK}
\end{document}